\documentclass[11pt]{article}

\usepackage[pdftex]{graphicx}
\usepackage[small,bf,sf,textfont={small,sf}]{caption}
\usepackage{vmargin}
\usepackage{amsmath}
\usepackage{amssymb}
\usepackage{fancyhdr}
\usepackage{color}
\usepackage[colorlinks=true,
            linkcolor=blue,
            urlcolor=blue,
            citecolor=blue]{hyperref}
\usepackage[round]{natbib}
\usepackage{doi}

\usepackage{booktabs}
\usepackage{float}
\usepackage{lineno}
\usepackage{epstopdf}
\usepackage{bm}
\usepackage{dsfont}
\usepackage{multirow}
\usepackage{pdflscape}
\usepackage{caption}
\usepackage{subcaption}
\usepackage{cprotect}
\usepackage{url}

\floatstyle{ruled}
\newfloat{floatbox}{thp}{lob}
\floatname{floatbox}{Algorithm}

\usepackage{titlesec}
\titleformat{\section}
   {\normalfont\sffamily\Large\bfseries}
   {\thesection}{1em}{}
\titleformat{\subsection}
   {\normalfont\sffamily\large\bfseries}
   {\thesubsection}{1em}{}
\titleformat{\subsubsection}
   {\normalfont\sffamily\bfseries}
   {\thesubsubsection}{1em}{}

\newcommand{\trp}{^{\!\top}}

\newcommand{\inv}{^{-1}}

\newcommand{\x}{\mathbf{x}}

\newcommand{\y}{\mathbf{y}}

\newcommand{\m}{\mathbf{m}}

\newcommand{\g}{\mathbf{g}}

\newcommand{\dsim}{\mathbf{d}}
\newcommand{\dobs}{\mathbf{d}_{\textrm{obs}}}

\newcommand{\A}{\mathbf{A}}

\newcommand{\C}{\mathbf{C}}

\newcommand{\Y}{\mathbf{Y}}

\newcommand{\Ce}{\mathbf{C}_{\mathbf{e}}}

\newcommand{\X}{\mathbf{X}}

\newcommand{\R}{\mathbf{R}}

\begin{document}


\title{\textbf{Mitigating Loss of Variance in Ensemble Data Assimilation: Machine Learning-Based Distance-Free Localization}}

\author{Vinicius L. S. Silva$^{a,b,*}$ \and Gabriel S. Seabra$^{a,c}$ \and Alexandre A. Emerick$^{a}$ }

\date{\begin{center}
\small \noindent $^a$Petróleo Brasileiro S.A. (Petrobras), Rio de Janeiro, Brazil \\
\small \noindent $^b$Novel Reservoir Modelling and Simulation Group, Imperial College London, UK\\
\small \noindent $^c$Faculty of Civil Engineering and Geosciences, TU Delft\\
\small \noindent $^*$ Corresponding author email: v.santos-silva19@alumni.imperial.ac.uk
\end{center}}

\maketitle


\begin{abstract}

Ensemble-based methods are among the most robust and efficient techniques for assimilating dynamic data in geological reservoir models, with applications in hydrocarbon production, groundwater management, geological carbon storage, and geothermal energy. In practice, however, the high computational cost of large-scale reservoir simulations limits the size of the ensemble, leading to sampling errors that can degrade assimilation performance. Localization is the standard approach to mitigate the adverse effects of small ensemble sizes. In its simplest form, localization is applied by tapering model updates based on the distance between observations and model parameters. However, selecting an appropriate localization region can be challenging and problem-specific. Furthermore, there are cases in which the model parameters and observations lack a clear spatial relationship, rendering distance-based localization inapplicable. To address these limitations, several distance-free localization methods have been proposed in the literature, but none has yet achieved widespread adoption. In this work, we proposed a novel distance-free localization strategy based on machine learning methods specifically tailored for tabular data. Additionally, we introduced a simple correction to the estimated prior cross-covariance, which can enhance localization values in problems involving a small number of model parameters. The proposed methods were integrated into the ensemble smoother with multiple data assimilation and tested on data assimilation problems with both scalar and grid-based model parameters. The results demonstrated that the machine learning-based localization improves cross-covariance estimates and, consequently, enhances data assimilation performance by producing posterior ensembles with higher variance while maintaining data match quality.
\\
\\
\textbf{Keywords:} 	ensemble data assimilation, history matching, localization, reservoir simulation, machine learning, covariance estimation
\end{abstract}

\section{Introduction}
\label{sec:intro}

Simulation models play a central role in the development and management of geological reservoirs across a range of applications, including hydrocarbon production, groundwater management, geological carbon storage, and geothermal energy. However, these models are inherently uncertain, which requires the assimilation of dynamic data collected during reservoir operations to reduce uncertainty and improve predictive reliability.

Among the most efficient and robust data assimilation techniques available today are iterative ensemble smoothers \citep{chen:13a,emerick:13b,luo:15b,silva:17a,raanes:19b}. These methods employ ensembles of model realizations to estimate covariances between uncertain model parameters and predicted data, which are then used to compute parameter updates that improve agreement with observations. Ensemble-based approaches naturally accommodate nonlinearities without requiring derivative computations, facilitating integration with complex forward models while maintaining feasible computational costs. Moreover, their Bayesian formulation allows for the generation of posterior ensembles that provide valuable uncertainty quantification for future predictions.

In practice, however, the high computational cost of large-scale reservoir simulations limits ensemble sizes, which introduces sampling errors manifested as spurious correlations between model parameters and predicted data in the estimated covariances. These spurious correlations can lead to erroneous updates, often resulting in excessive variance reduction or even ensemble collapse, ultimately compromising the effectiveness of the data assimilation process.

To mitigate variance underestimation caused by spurious correlations, localization is widely employed \citep{houtekamer:01}. Traditional localization techniques assume that covariances between model parameters and data decay with physical distance. The most common implementation involves the Schur product between the estimated covariance (or Kalman gain) and a compactly supported correlation matrix, effectively tapering the influence of each component of the innovation vector (data mismatch between observations and predictions). Distance-based localization methods are simple, computationally efficient, and often highly effective \citep[Chap.~7]{emerick:25bk}. Nonetheless, defining an appropriate localization region is problem-specific and requires careful consideration of reservoir dynamics and data distribution.

Importantly, many of the key parameters used in reservoir data assimilation, such as multipliers for rock properties, relative permeability curves, analytical aquifers, and compressibilities are not defined on spatial grids and do not have a straightforward spatial relationship with observational data. These scalar parameters often exert a regional or even global influence on the behavior of the system. In such cases, conventional distance-based localization fails, and alternative methods that do not rely on the spatial proximity are required.

Several approaches have been proposed for constructing distance-free localization matrices that avoid the assumption of spatial correlation decay \citep{furrer:07,anderson:07,bishop:09a,zhang:10,luo:18a,vishny:24a}. \citet{lacerda:19a} reviewed and compared several of these methods in the context of updating scalar parameters in a modified version of the PUNQ-S3 benchmark model \citep{floris:01}. Their findings showed that none of the tested methods were entirely successful in avoiding variance underestimation, and in some cases, they even impaired the ability to match the data. Most approaches also required significant tuning to achieve reasonable performance.

To address these limitations, \citet{lacerda:21a} proposed using least-squares support vector regression \citep{suykens:99a} combined with the tapering strategy from \citet{furrer:07} to estimate localization coefficients. When applied to the same modified PUNQ-S3 case, this approach led to small/moderate improvements in posterior ensemble variance estimation.

In this work, we extend the approach proposed by \citet{lacerda:21a} by exploring state-of-the-art machine learning (ML) models tailored for tabular data. Additionally, we introduce a simple analytical correction to the estimated prior cross-covariance that can enhance the localization values. Our goal is to develop a robust, distance-free localization strategy that improves covariance estimation and mitigates the loss of variance in ensemble data assimilation. We first apply the proposed approaches to scalar parameters and then extend the investigation to grid-based parameters. We conduct the experiments on the same PUNQ-S3 case study used in \citet{lacerda:19a,lacerda:21a} to enable direct comparison and evaluation of improvements, and on a geological carbon storage case first introduced by \citet{seabra:24a}.  

The remainder of this paper is organized as follows. Section~\ref{sec:ensemble} provides a brief overview of ensemble-based data assimilation, with an emphasis on the Ensemble Smoother with Multiple Data Assimilation (ES-MDA) \citep{emerick:13b}, which serves as the foundation for all numerical experiments. This section also reviews the implementation of localization using the taper function proposed by \citet{furrer:07}. Section~\ref{sec:method} describes the proposed methods, while Section~\ref{sec:test_cases} presents results for test cases involving both scalar and grid-based model parameters. Section~\ref{sec:disc} discusses the performance and potential limitations of the proposed methods. Finally, Section~\ref{sec:conc} summarizes the main conclusions.

\section{Ensemble Based-Data Assimilation}
\label{sec:ensemble}

Ensemble-based data assimilation methods have become widely used for integrating observational data into numerical models while accounting for uncertainty in both model parameters and measurements. These methods rely on an ensemble of model realizations to estimate the covariances required for updating model parameters, offering a practical framework for solving high-dimensional and nonlinear inverse problems. Among these methods, iterative ensemble smoothers are the preferred choice for reservoir applications, as they allow the assimilation of all available data in a computationally efficient manner and avoid the need for forward model linearization or adjoint implementations \citep{emerick:25bk}. In particular, ES-MDA have demonstrated strong performance across a wide range of data assimilation problems. ES-MDA perform successive global updates of the ensemble by applying a sequence of smaller corrections, which enhances robustness in nonlinear settings and improves the consistency between model parameters and predicted responses. 

In this work, we employ ES-MDA as the primary data assimilation method, although the proposed localization approaches can be readily applied to other ensemble-based techniques. In the following sections, we briefly review the ES-MDA update equation and discuss localization schemes, including a distance-free localization strategy.

\subsection{Ensemble Smoother with Multiple Data Assimilation}
\label{sec:ensemble.esmda}

ES-MDA is an ensemble-based method developed for parameter estimation problems, in which multiple smaller updates are performedm mimicking an iterative process. The method is motivated by the equivalence between single and multiple data assimilation in linear-Gaussian systems \citep{emerick:13b}, where the same data are assimilated multiple times using an inflated measurement error covariance matrix at each step.

For a vector of model parameters $\m \in \mathds{R}^{N_m}$, the ES-MDA analysis equation can be written as:

\begin{equation}\label{eq:ensemble.esmda}
  \m^{k+1}_j = \m^k_j + \widetilde{\C}^k_{\m\dsim} \left(\widetilde{\C}^k_{\dsim\dsim} + \alpha_k\Ce \right)\inv \left(\dobs + \mathbf{e}^k_j - \dsim^k_j \right),
\end{equation}

\noindent for $j = 1,\ldots,N_e$, where $N_e$ denotes the ensemble size. In this equation, $\dobs \in \mathds{R}^{N_d}$ is the vector of observed data, $\mathbf{e}^k_j \in \mathds{R}^{N_d}$ is the vector of random perturbations sampled from $\mathcal{N}(0,\alpha_k\Ce)$, with $\Ce \in \mathds{R}^{N_d \times N_d}$ denoting the data-error covariance matrix. $\dsim^k_j \in \mathds{R}^{N_d}$ is the vector of predicted data, and $N_d$ is the number of data points. The matrices $\widetilde{\C}^k_{\m\dsim} \in \mathds{R}^{N_m \times N_d}$ and $\widetilde{\C}^k_{\dsim\dsim} \in \mathds{R}^{N_d \times N_d}$ are the cross-covariance between model parameters and predicted data and the auto-covariance matrix of predicted data, respectively. The tilde over these matrices indicate that they are estimated using the current ensemble.

The inflation coefficients $\alpha_k$ must be selected in advance such that:

\begin{equation}\label{eq:ensemble.esmda.alpha}
    \sum_{k=1}^{N_a} \frac{1}{\alpha_k} = 1,
\end{equation}

\noindent where $N_a$ is the number of data assimilations. A common choice is to use constant coefficients, i.e., $\alpha_k = N_a$ for all $k$ \citep{emerick:13b}. Although other choices are possible, experimental results indicate that using constant coefficients often provides satisfactory results \citep{emerick:16a}.

\subsection{Localization}
\label{sec:method.local}

In practical applications of ensemble-based methods, the limited ensemble size leads to sampling errors in the estimation of covariance matrices. These errors introduce spurious correlations between model parameters and predicted data, which can result in overly aggressive or inaccurate updates. Consequently, the ensemble may undergo severe variance reduction or, in extreme cases, collapse entirely, compromising its ability to effectively assimilate new data \citep{hamill:01,aanonsen:09}.

The standard approach to mitigate these sampling errors is localization \citep{houtekamer:01}. In practice, localization is applied by making the Schur (Hadamard) product between the localization matrix $\R \in \mathds{R}^{N_m \times N_d}$ and the Kalman gain matrix \citep{emerick:16a}. This leads to a modified ES-MDA analysis equation:

\begin{equation}\label{eq:ensemble.local.esmda}
    \m^{k+1}_j = \m^k_j + \R \circ \left[ \widetilde{\C}^k_{\m\dsim} \left(\widetilde{\C}^k_{\dsim\dsim} + \alpha_k\Ce \right)\inv \right] \left(\dobs + \mathbf{e}^k_j - \dsim^k_j \right)
\end{equation}

\noindent where ``$\circ$'' denotes the Schur product. 

In distance-based localization, the $(i,k)$th entry of the localization matrix $\R$, denoted as $r_{ik}$, is computed using a compactly supported correlation function that depends on the distance between the $i$th model parameter and the $k$th data point. In practical reservoir applications, the Gaspari-Cohn correlation function \citep{gaspari:99} is commonly used. This function requires the specification of a critical length, which defines the region around each data location within which updates are applied and beyond which the influence tapers to zero.

For parameters that do not have a direct spatial relationship with observations, alternative localization schemes are required. In this work, we adopt the pseudo-optimal taper function proposed by \citet{furrer:07} as the basis for the localization method developed herein. This taper function was chosen for its solid mathematical foundation and ease of implementation. Additionally, it was one of the best-performing methods in the evaluation by \citet{lacerda:19a}.

The pseudo-optimal localization scheme was derived by minimizing, term by term, the expected Frobenius norm of the difference between the true covariance matrix and its localized estimate. The resulting expression for the $(i,k)$th entry of the localization matrix $\mathbf{R}$ is given by

\begin{equation}\label{eq:ensemble.local.po}
    r_{ik} = \frac{c_{ik}^2}{c_{ik}^2 + \frac{c_{ik}^2 + c_{ii}c_{kk}}{N_e}},
\end{equation}

\noindent where $c_{ik}$ denotes the true covariance between the $i$th model parameter and the $k$th predicted data point. In practice, however, $c_{ik}$ is unknown. To address this, \citet{furrer:07} proposed replacing $c_{ik}$ with its corresponding ensemble estimate $\widetilde{c}_{ik}$, and setting small covariance values to zero, which can be done with a simple thresholding rule: $r_{ik} = 0$ if $|\widetilde{c}_{ik}| < \eta\sqrt{\widetilde{c}_{ii}\widetilde{c}_{kk}}$, where $\eta \in [0, 1)$ is a small user-defined threshold. In this text, we refer this method as PO-localization. The choice of $\eta$ influences the performance of the method: larger values lead to more aggressive removal of spurious covariances but may hinder the ability to properly assimilate data \citep{lacerda:19a}. In this work, we adopt a small threshold, $\eta = 10^{-3}$, as our goal is to investigate the effects of the proposed localization scheme without overly restricting the updates.

\section{Proposed Methods}
\label{sec:method}

In this section, we describe the proposed distance-free localization method, which combines ML with the PO taper function. Additionally, we introduce a simple analytical correction to improve the estimation of the prior cross-covariance, which is particularly useful for problems involving a small number of model parameters.

\subsection{Machine Learning Localization}
\label{sec:method.ml-local}

In this work, we extend the localization approach proposed in \citep{lacerda:21a} by leveraging state-of-the-art ML methods designed for tabular data. The method uses the prior ensemble of model parameters and their corresponding predicted data as training set for a ML proxy. Once trained, the proxy is used to estimate the cross-covariances between model parameters and predicted data using a large ensemble of size $N_E \gg N_e$, which is then employed to compute the localization coefficients via Eq.~\eqref{eq:ensemble.local.po}. The proposed method is presented in Algorithm~\ref{algo:method.ml-local}:

\begin{floatbox}{}{}
\begin{tabular}{ll}
    $\texttt{1.}$ & \verb"Compute predicted data with the prior ensemble" \\
    ~ & $\left\{ \m_j\right\}_{j=1}^{N_e} \rightarrow \left\{ \dsim_j = \g\left( \m_j \right)\right\}_{j=1}^{N_e}$ \\
    ~ & ~ \\
    
    $\texttt{2.}$ & \verb"Train a ML proxy using the prior ensemble"\\
    ~ & $\widehat{\dsim} = \mathcal{G}\left(\m\right)$ \\
    ~ & ~ \\
    
    $\texttt{3.}$ & \verb"Compute predicted data with the ML proxy in a large" \\
    $~$ & \verb"ensemble" $(N_E \gg N_e)$ \\
    ~ & $\left\{ \m_j\right\}_{j=1}^{N_E} \rightarrow \left\{ \dsim_j = \mathcal{G}\left( \m_j \right)\right\}_{j=1}^{N_E}$\\
    ~ & ~ \\
    
    $\texttt{4.}$ & \verb"Compute cross-covariances using the large ensemble and the" \\ 
    $~$ & \verb"data predicted by the ML proxy"\\
    ~ & $\widehat{\C}_{\m\dsim} = \frac{1}{N_E - 1} \sum_{j=1}^{N_E} \left(\m_j - \overline{\m} \right)\left(\widehat{\dsim}_j - \overline{\widehat{\dsim}} \right)\trp$\\
    ~ & ~ \\
    
    $\texttt{5.}$ & \verb"Compute the localization coefficients"\\
    ~ &  $r_{ik} = \frac{\widehat{c}_{ik}^2}{\widehat{c}_{ik}^2 + \left(\widehat{c}_{ik}^2 + \widehat{c}_{ii}\widehat{c}_{kk}\right)/N_e}$, \verb"for" $i \in [1,N_m]$ \verb"and" $k \in [1,N_d]$ \\
    ~ &  \verb"Set" $r_{ik} = 0$ \verb"if" $|\widehat{c}_{ik}| < \eta\sqrt{\widehat{c}_{ii}\widehat{c}_{kk}}$ \\
\end{tabular}
\caption{ML-localization}\label{algo:method.ml-local}
\end{floatbox}

The first step of Algorithm~\ref{algo:method.ml-local} consists of selecting the prior ensemble of model parameters and corresponding data predictions, $\{\m_j, \dsim_j\}_{j=1}^{N_e}$, as the training dataset for the ML proxy. This is the same ensemble used during the data assimilation process, meaning that no additional reservoir simulations are required. Furthermore, we compute the localization coefficients using only the prior ensemble. Results presented in \citet{lacerda:19a} indicate that this strategy is more robust than updating the localization coefficients at each iteration of the ES-MDA method. We tested alternative strategies and the results are presented in Section~\ref{sec:test_cases.scalar.sens}.

In the second step of Algorithm~\ref{algo:method.ml-local}, an ML proxy, denoted by $\mathcal{G}(\cdot)$, is trained to predict data responses from model parameters. This proxy is then used to generate predicted data for a large ensemble of size $N_E \gg N_e$, i.e., $\widehat{\dsim}_j = \mathcal{G}(\m_j)$ for $j = 1, \dots, N_E$ (step 3). This large ensemble is used to estimate the cross-covariance matrix between model parameters and data, denoted by $\widehat{\C}_{\m\dsim}$ (step 4). The entries of this matrix are then used to compute the localization coefficients in step 5 using the PO taper.

The rationale behind this approach is that the cross-covariance $\widehat{\C}_{\m\dsim}$, estimated from a large ensemble generated via the ML proxy, provides a more accurate approximation of the true cross-covariance $\C_{\m\dsim}$ than $\widetilde{\C}_{\m\dsim}$, which is computed from data assimilation ensemble with $N_e$ members. Note that the symbols $~\widehat{ }~$ and $~\widetilde{ }~$ represent the estimated cross-covariance calculated using the large ($N_E$) and data assimilation ($N_e$) ensembles, respectively. It is also important to note that although the localization coefficients are computed using an ensemble of $N_E$ members, we use $N_e$ in the expression for the PO taper (step 5), since these coefficients are applied to localize the updates in the data assimilation ensemble. In all examples discussed in the paper, we set $N_E = 5{,}000$. 

Although we use Eq.~\eqref{eq:ensemble.local.po} to compute the localization coefficients (step 5), it is worth noting that the proposed scheme can be applied with other taper functions as well. For example, we tested the method using the correlation-based localization scheme described in \citep{luo:18b,luo:20a} and observed improvements consistent with those reported in this paper using the PO taper. In the following, we refer to the machine learning-based localization proposed in this section as ML-localization.

\subsubsection{Machine Learning Methods}
\label{sec:method.ml-local.ml-methods}

The number of ML methods available in the literature is extensive. ML has become one of the most active areas of research and development, with a wide range of applications across diverse fields. This rapid growth has been driven in part by the availability of efficient and accessible implementations of widely used ML algorithms. 

Recent advances in deep learning, particularly attention-based architectures such as TabNet \citep{arik21:a} and FT-Transformer \citep{gorishniy:21a}, have shown promising results on tabular datasets. Tabular data, structured in rows and columns with each row representing a sample and each column a feature, is common in reservoir simulations and data assimilation workflows. Nonetheless, despite the remarkable progress of deep learning in recent decades, studies have shown that traditional, or ``classical'', ML methods, based on ensemble of decision trees, often outperform deep learning techniques when applied to tabular data \citep{shwartzziv:22a,borisov:24a,grinsztajn:22a}. 


Given the variety of available methods, we compared the performance of ten different ML algorithms in the first part of Section~\ref{sec:test_cases}, including ``classical'' ML and deep learning methods, with the goal of identifying the most suitable approach for generating localization coefficients. The following methods were considered:

\begin{itemize}
    \item \textbf{Linear regression}: Linear regression predicts output by computing a weighted sum of the input features plus a constant. The output is a linear function of the inputs. We use the scikit-learn implementation of linear regression \citep{scikit-learn:11a}.

    \item \textbf{Decision tree}: Decision trees have a hierarchical structure consisting of nodes and branches. They partition the feature space into regions and assign a constant prediction within each region. Although simple, they are powerful and serve as the foundation for ensemble methods like random forests and gradient boosting \citep{hastie:09a,geron:19a}. We use the scikit-learn implementation.

    \item \textbf{Support vector regression (SVR)}: SVR is based on the principles of Support Vector Machines and aims to fit the best hyperplane to the training data within a predefined margin of tolerance. It can model both linear and nonlinear relationships through kernel functions \citep{smola:04a}. We use the scikit-learn implementation with a radial basis function (RBF) kernel.

    \item \textbf{Random forest}: Random forests \citep{breiman:01a} are ensembles of decision trees trained on bootstrap samples of the data. At each node, a random subset of features is considered for splitting, which reduces variance and improves generalization \citep{breiman:98a}. The final prediction is obtained by averaging the outputs of all trees. We use the scikit-learn implementation.

    \item \textbf{Extra trees}: Extremely randomized trees (Extra Trees) are similar to random forests but introduce additional randomness by selecting split thresholds at random \citep{geurts:06a}. This can reduce variance at the cost of slightly increased bias. We use the scikit-learn implementation.

    \item \textbf{XGBoost}: Extreme Gradient Boosting (XGBoost) is an advanced gradient-boosted decision tree (GBDT) algorithm \citep{chen:16c}. It builds trees sequentially, with each tree correcting the residuals of its predecessor. XGBoost introduces regularization and other enhancements, achieving state-of-the-art performance on many tabular datasets \citep{shwartzziv:22a,borisov:24a,grinsztajn:22a}. We use the official XGBoost implementation \citep{xgboost}.

    \item \textbf{LightGBM}: LightGBM is a GBDT algorithm designed for speed and scalability. It incorporates Gradient-based One-Side Sampling and Exclusive Feature Bundling to efficiently handle large datasets and high-dimensional features \citep{ke:17a}. We use the official LightGBM implementation \citep{lightgbm} combined with multi-target regression of scikit-learn since LightGBM does not natively support multiple outputs.  

    \item \textbf{Multilayer perceptron (MLP)}: An MLP is a fully connected feedforward neural network composed of at least three layers: input, hidden, and output. Each neuron applies a nonlinear activation function, enabling the network to learn complex patterns \citep{goodfellow:16bk,hastie:09a}. We implement the MLP using TensorFlow \citep{tensorflow:15}.

    \item \textbf{TabNet}: TabNet is a deep learning architecture for tabular data that uses sequential attention to select relevant features at each decision step \citep{arik21:a}. Feature selection is performed on a per-instance basis, allowing the model to adaptively focus on different features for different inputs. We use the PyTorch implementation \citep{pytorch}.

    \item \textbf{FT-Transformer}: The FT-Transformer is an adaptation of the transformer architecture \citep{vaswani:17a} for tabular data, proposed by \citet{gorishniy:21a}. It applies embeddings to both numerical and categorical features, followed by a stack of transformer layers with parallel attention. We use the PyTorch implementation.
\end{itemize}

We chose not to perform hyperparameter optimization for the ML algorithms. Instead, we use the default hyperparameter values provided by each implementation to avoid the computational cost of tuning and to promote a more robust and practical approach that does not rely on model-specific calibration.

\subsection{Correction of the Prior Covariance}
\label{sec:method.cm-local}

Here, we propose a straightforward correction to the prior cross-covariance that can be used to improve the calculation of localization coefficients. This correction is inspired by the application of linear regression as a ML proxy, as discussed in the previous section. Given the training set $\{\m_j, \dsim_j\}_{j=1}^{N_e}$, the linear regression model can be expressed as

\begin{equation}\label{eq:method.cm-local.lr}
    \Y = \X \A,
\end{equation}

\noindent where $\Y \in \mathds{R}^{N_e \times N_d}$ is the matrix whose $j$th row corresponds to the predicted data of the $j$th ensemble member, $\y_j\trp = \dsim_j\trp$; $\X \in \mathds{R}^{N_e \times (N_m + 1)}$ is the design matrix whose $j$th row is $\x_j\trp = \left[1,~\m_j\trp \right]$, including a bias term; and $\A \in \mathds{R}^{(N_m + 1) \times N_d}$ is the matrix of regression coefficients. The coefficient matrix $\A$ is obtained by solving a least-squares problem, yielding

\begin{equation}\label{eq:method.cm-local.A}
    \A =  \left( \X\trp \X \right)\inv \X\trp \Y.
\end{equation}

The cross-covariance between $\x$ and $\y$ can be estimated from the training set using Eq.~\eqref{eq:method.cm-local.lr} and \eqref{eq:method.cm-local.A} as

\begin{equation}\label{eq:method.cm-local.A}
   \widetilde{\C}_{\x\y} = \frac{1}{N_e -1} \X\trp\Y,
\end{equation}

\noindent Here, in order to simplify the notation, we assume that the matrices $\X$ and $\Y$ are already centered (i.e., their row means have been subtracted). However, it is worth noting that the final conclusion would be the same if the equations were written in terms of $\Delta \X = \X - \overline{\X}$ and $\Delta \Y = \Y - \overline{\Y}$, where $\overline{\X}$ and $\overline{\Y}$ are matrices in which all rows are equal to the corresponding mean vectors. Also note that we use the transpose matrices $\X\trp$ and $\Y\trp$ to calculate the estimate covariance ($\X\trp\Y$ instead of $\X\Y\trp$) because the vectors $\x$ and $\y$ are represented as rows in $\X$ and $\Y$, instead of columns. 

In the proposed ML localization procedure we use a large ensemble with $N_E$ elements to estimate the cross-covariance employed to compute the localization coefficients. Following the same procedure, we write

\begin{equation}\label{eq:method.cm-local.Yhat}
    \widehat{\Y} = \widehat{\X} \A = \widehat{\X}\left( \X\trp \X \right)\inv \X\trp \Y,
\end{equation}

\noindent where the hat over the matrices indicate that they correspond to the large ensemble ($N_E$ rows). The cross-covariance can be estimated as
\begin{eqnarray}\label{eq:method.cm-local.Chat1}
   \nonumber \widehat{\C}_{\x\y} & = & \frac{1}{N_E - 1} \widehat{\X}\trp\widehat{\Y} \\
   \nonumber & = & \frac{1}{N_E - 1} \widehat{\X}\trp \widehat{\X}\left( \X\trp \X \right)\inv \X\trp \Y \\
    & = & \widehat{\C}_{\x\x} \widetilde{\C}_{\x\x}\inv \widetilde{\C}_{\x\y}. 
\end{eqnarray}

This result shows that using linear regression to generate predictions for a large ensemble in order to estimate the cross-covariance can be interpreted as correcting the covariance $\widetilde{\C}_{\x\y}$ by replacing the auto-covariance estimate $\widetilde{\C}_{\x\x}$ with $\widehat{\C}_{\x\x}$. Eq.~\eqref{eq:method.cm-local.Chat1} can be written in terms of the vectors of model parameters and data by simply replacing, $\x$ by $\m$ and $\y$ by $\dsim$. Moreover, the true prior covariance $\C_{\m\m}$ is often known, particularly in the case of scalar model parameters, where the prior is typically prescribed and frequently assumed to be diagonal (which is the case for independent scalar parameters). In such cases, rather than relying on a linear proxy, we can directly write the corrected cross-covariance estimate as

\begin{equation}\label{eq:method.cm-local.Chat2}
    \widehat{\C}_{\m\dsim} = \C_{\m\m} \widetilde{\C}_{\m\m}\inv \widetilde{\C}_{\m\dsim}.
\end{equation}

Eq.~\eqref{eq:method.cm-local.Chat2} represents an improved estimate of the prior cross-covariance matrix. We propose using $\widehat{\C}_{\m\dsim}$ from this expression to compute the localization coefficients using the PO taper (Eq.~\ref{eq:ensemble.local.po}). The inverse $\widetilde{\C}_{\m\m}\inv$ in Eq.~\eqref{eq:method.cm-local.Chat2} is replaced by the pseudo-inverse computed with singular value decomposition. In the following, we refer to this procedure as CM-localization. This correction can also be applied when the cross-covariances are estimated using the ML proxy. Conceptually, the prior covariance correction can always be used as long as it is feasible to compute and store the full $\C_{\m\m}$ matrix. In practice, however, this approach is only feasible in cases with a small number of model parameters. For this reason, we propose using it to compute localization coefficients for scalar model parameters, which, in typical reservoir applications, are usually limited to a few hundred.

\section{Results}
\label{sec:test_cases}

In this section, we evaluate the performance of the proposed localization strategies. We begin by assessing the ML- and CM-localization approaches in a modified version of the PUNQ-S3 problem that includes only scalar model parameters. In this case, we compare the performance of different ML methods in improving the estimates of the cross-covariance between model parameters and predicted data. Next, we apply the ML-localization to two problems involving grid-based parameters. The first is a small-scale CO$_\text{2}$ injection problem designed to mimic a carbon capture and storage (CCS) scenario. The second is the PUNQ-S3 case, in which we update grid properties corresponding to porosity, as well as horizontal and vertical permeability. Finally, we present the results for the computational time required to train the ML proxies.

\subsection{Scalar Parameters}
\label{sec:test_cases.scalar}

\begin{figure}[!htb]
	\centering
	\includegraphics[width=0.5\textwidth]{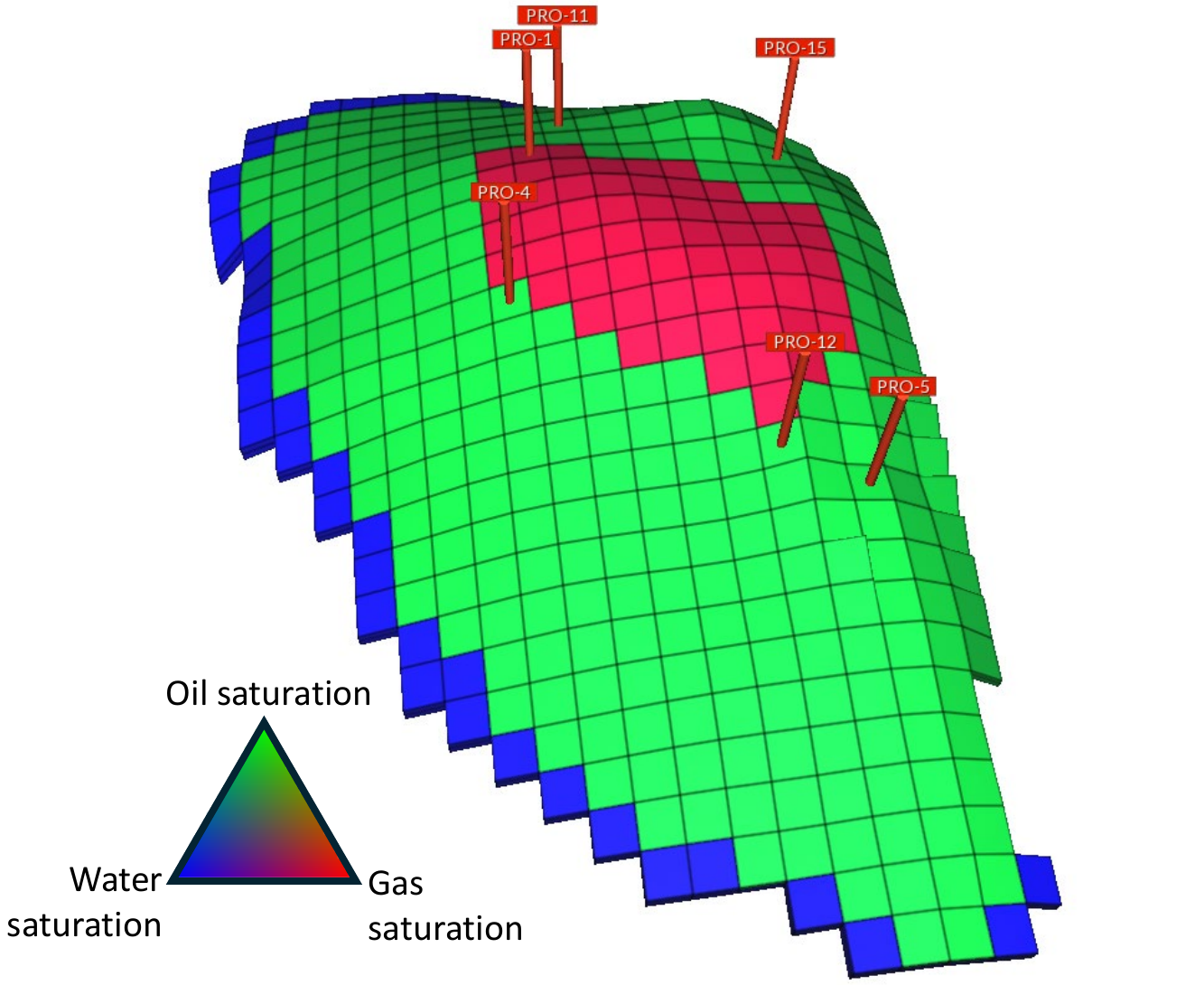}
	\caption{PUNQ-S3 model.}
	\label{fig:test_cases.scalar.punq}
\end{figure}

The PUNQ-S3 reservoir model is a benchmark case developed to evaluate the performance of history matching methods, particularly in terms of uncertainty quantification. The model is based on a real field from Elf Exploration Production and represents a three-phase, three-dimensional porous media flow system. It employs a corner-point grid with 19~$\times$~28~$\times$~5 cells, totaling 2,660 gridblocks, of which 1,761 are active. The gridblocks have uniform horizontal dimensions of 180 meters in both length and width, but variable thickness. The geological structure features sand channels embedded in low-porosity shale, with facies variations across layers. The model simulates multiphase flow involving compressible gas, oil, and water, making it well-suited for testing advanced data assimilation techniques. Further details about the PUNQ-S3 model can be found in \citet{floris:01}. Fig.~\ref{fig:test_cases.scalar.punq} illustrates the PUNQ-S3 model by showing the initial distribution of gas, oil, and water. The reservoir is connected to a large aquifer, which is represented using analytical model.

Here, we consider a modified version of the PUNQ-S3 case, as discussed in \citet{lacerda:19a}, which includes only scalar model parameters. A commercial black-oil simulator is used to perform the reservoir simulations.  During the historical period, six oil-producing wells operate under oil-rate control. The production history consists of an initial year of extended well testing, followed by a three-year shut-in period, and then four additional years of production. We consider 15 scalar model parameters, which includes five porosity multipliers (one per reservoir layer), five permeability multipliers (one per reservoir layer), two analytical aquifer radii, the water-oil and gas-oil contacts, and the rock compressibility. The prior ensemble was generated by sampling each parameter independently, resulting in a diagonal prior covariance matrix.

In addition to the model parameters, we introduced five dummy parameters (which have no influence on the predicted data) to evaluate the effect of localization on uncorrelated variables. Since these dummy parameters do not affect the model predictions, any updates to their values are solely due to sampling errors.

The observed data include measurements of well water cut (WWCT), well gas-oil ratio (WGOR), and well bottom-hole pressure (WBHP), which were generated by adding random noise to the data predicted by the ``ground truth'' model. The noise was sampled from a Gaussian distribution with zero mean and a standard deviation equal to 10\% of the data value for WWCT and WGOR, and 1\% for WBHP. The total number of data points is 1,530.

\subsubsection{Cross-Correlation Estimation}
\label{sec:test_cases.scalar.corr_est}

In our first set of tests, we compare the effectiveness of different localization strategies in improving the estimation of cross-correlations between model parameters and predicted data. Correlations are used instead of covariances to mitigate the impact of differing magnitudes between model parameters and predicted data when computing comparison metrics. For this purpose, we define a ``gold standard'' as the cross-correlation computed from an ensemble of 5,000 realizations without applying any localization strategy. Table~\ref{tab:test_cases.scalar.corr_est} summarizes the results for an ensemble size of $N_e = 200$. We compare results obtained using no localization, PO-localization, CM-localization, and ML-localization. For the ML-based approach, we evaluate all ten selected machine learning methods and include columns reporting both the training time and the root mean square error (RMSE) on a test set consisting of 1,000 additional realizations that were not used for training. In the ML-localization approach, an ensemble of $N_E = 5{,}000$ realizations is used to compute the localization matrix. To facilitate visual comparison, the table cells are color-coded using a red-yellow-green scale, with green indicating better performance. As evaluation metrics, we use two matrix norms to quantify the difference between the estimated cross-correlation matrices and the gold standard: the Frobenius norm, which corresponds to the element-wise RMSE, and the spectral norm, which reflects the largest singular value of the difference matrix.

The results in Table~\ref{tab:test_cases.scalar.corr_est} show that the case without localization yielded the largest values for both the Frobenius and spectral norms. PO-localization provided only a modest improvement, whereas the introduction of the proposed prior correction (CM-localization) led to a significant enhancement in accuracy. Most of the ML-localization methods also produced notable improvements in the matrix norms, with the exception of Decision Tree and TabNet, which exhibited the highest RMSE values on the test set. The lowest matrix norm was achieved using ML-localization with the FT-Transformer; however, this method also had the highest training time. It is important to note that the current test problem is relatively small compared to typical operational models, as it includes only 20 parameters and 1,530 data points. Consequently, training time may become a limiting factor for the application of such methods in real-world settings. Among the tested algorithms, the gradient boosting decision tree methods (XGBoost and LightGBM) offered a favorable balance between training time, test set RMSE, and matrix norm accuracy.

\begin{table}
	\centering
    \caption{Comparison of different localization methods in terms of cross-correlation estimation error. Table cells are color-coded using a red-yellow-green scale, with green indicating better performance.}
	\includegraphics[width=1.0\textwidth]{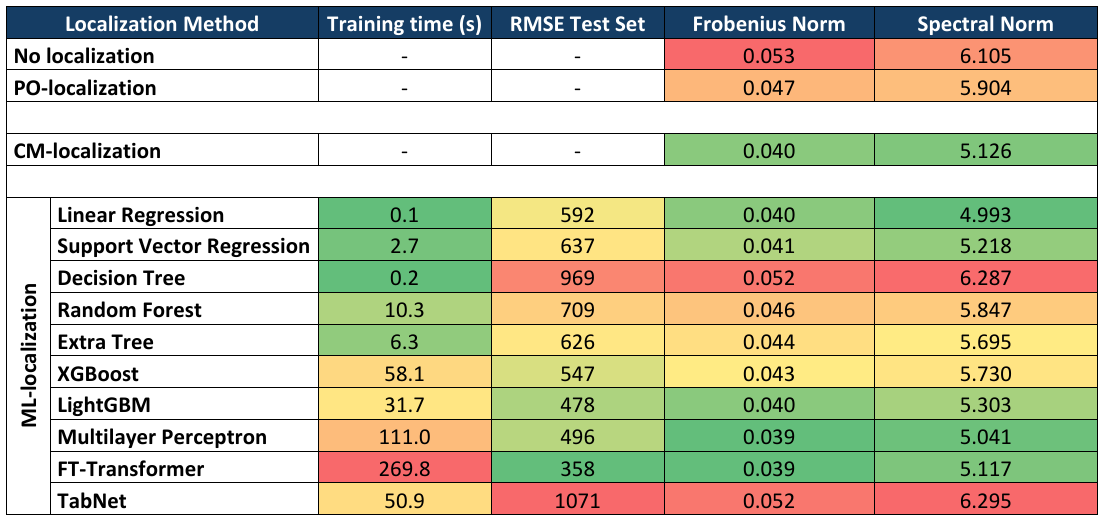}
	\label{tab:test_cases.scalar.corr_est}
\end{table}

\begin{figure}
	\centering
	\includegraphics[width=\textwidth]{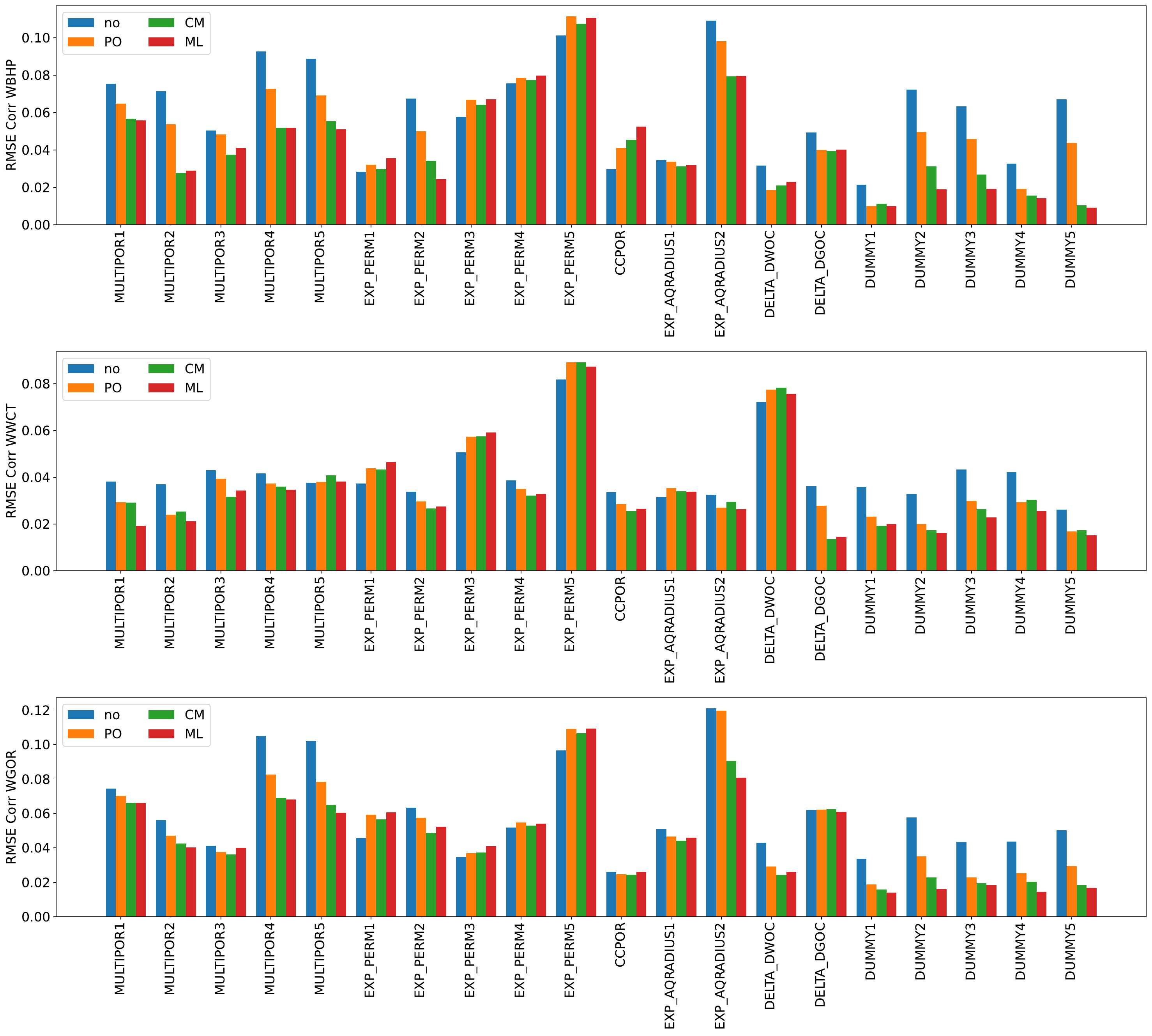}
	\caption{Cross-correlation RMSE for different model parameters grouped by data type. Blue bars represent the case with no localization, orange bars correspond to PO-localization, green bars to CM-localization, and red bars to ML-localization using LightGBM.}
	\label{fig:test_cases.scalar.rmse_data}
\end{figure}

Fig.~\ref{fig:test_cases.scalar.rmse_data} expands on the results presented in Table~\ref{tab:test_cases.scalar.corr_est} by showing the cross-correlation RMSE for individual model parameters, grouped by data type. The results show that ML-localization (using LightGBM) yields the lowest error for most parameters, followed by CM-localization, which consistently produces the second lowest errors. Additionally, for the dummy variables, where no correlation is expected, both ML- and CM-localization significantly reduce the error compared to PO-localization and the case without localization.

\begin{figure}
    	\centering
    	\includegraphics[width=1.0\columnwidth]{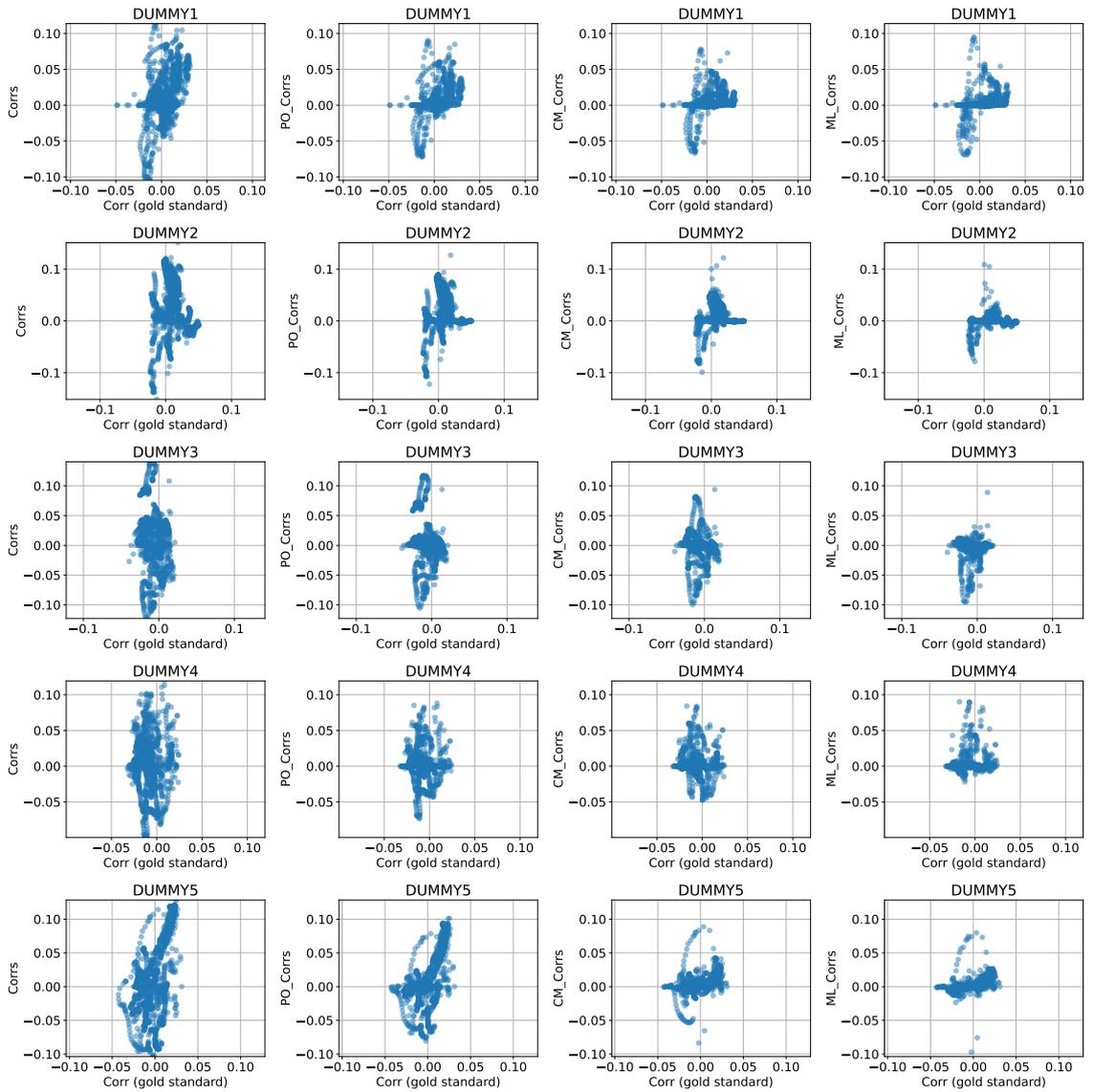}
        \caption{Comparison of cross-correlation values obtained using no localization (Corrs), PO-localization (PO\_Corrs), CM-localization (CM\_Corrs), and ML-localization with LightGBM (ML\_Corrs). The horizontal axes show the ``gold standard'' correlations, while the vertical axes show the corresponding localized correlations. Each row corresponds to a different dummy variable, and each column represents a different localization scheme.}
        \label{fig:test_cases.scalar.corr_xplot}
\end{figure}

Fig.~\ref{fig:test_cases.scalar.corr_xplot} compares the cross-correlation values of the dummy parameters obtained using no localization, PO-localization, CM-localization, and ML-localization (with LightGBM). In all plots, the horizontal axis represents the gold standard correlations, while the vertical axis shows the localized correlations. Each blue dot corresponds to the correlation coefficient between a dummy parameter and an observed data point. Since the dummy parameters are uncorrelated with the predicted data and have independent priors, the true correlation values should be zero. Notably, even with a large ensemble, the estimated cross-correlations (horizontal axis) for the dummy variables deviate from zero, indicating that eliminating sampling error would require an ensemble size much larger than $N_E = 5{,}000$. The plots demonstrate that both CM-localization and ML-localization substantially reduce the vertical spread of the points, indicating improved cross-correlation estimates compared to PO-localization and the case without localization.

\subsubsection{Data Assimilation Results}
\label{sec:test_cases.scalar.data_assim}

In this section, we evaluate the performance of the methods in terms of data assimilation results. We perform the data assimilation using the ES-MDA with four iterations with no localization, PO-localization, CM-localization, and ML-localization. For ML-localization, we consider only the LightGBM algorithm. Table~\ref{tab:test_cases.scalar.data_assim} summarizes the results for data assimilation using three ensemble sizes: 50, 100, and 200. For comparison, we also include the results of a data assimilation run without localization using an ensemble of 5,000 realizations. We use the following metrics to evaluate the quality of the data assimilation:

\begin{itemize}
    \item Mean data mismatch objective function: The objective function for each model in the ensemble is calculated as 
        \begin{equation}
            O(\m_j) = \frac{1}{2N_d}\left(\dobs - \dsim_j \right)\trp\Ce\inv\left(\dobs - \dsim_j \right),
        \end{equation}
        for $j = 1, 2, \ldots, N_e$. The objective function quantifies the quality of the data match. 
    \item Normalized variance (NV): The NV of each ensemble is computed as
        \begin{equation}
            \text{NV} = \frac{1}{N_m} \sum_{i=1}^{N_m} \frac{\text{var}[m'_i]}{\text{var}[m_i]},
        \end{equation}
        where $\text{var}[m'_i]$ and $\text{var}[m_i]$ denote the $i$th posterior and prior model parameters variance, respectively. This metric indicates how much of the prior variance is retained in the posterior ensemble. For dummy parameters, we expect the NV to be close to one, meaning no variance reduction due to data assimilation.
  
    \item Absolute mean offset (AMO) for dummy parameters:
        \begin{equation}
            \text{AMO} = \frac{1}{N_m} \sum_{i=1}^{N_m} |\text{mean}[m'_i] -\text{mean}[m_i]|,
        \end{equation}
        where $\text{mean}[m'_i]$ and $\text{mean}[m_i]$ denote the posterior and prior means for the $i$th model parameter, respectively.  This metric measures the shift in the mean caused by data assimilation. For dummy parameters, we expect the mean offset to be close to zero.
    \item Two statistical divergences for the dummy parameters:  We compute the Jensen–Shannon divergence (JS) \citep{lin:91a} and the Bhattacharyya coefficient (BC) \citep{cam:00a} to quantify the difference between prior and posterior distributions. Since dummy parameters have no influence on the predicted data, their posterior distributions should remain nearly identical to their priors, resulting in divergence values close to zero.
\end{itemize}

\begin{table}
	\centering
    \caption{Comparison of different localization methods in terms of data assimilation performance. Table cells are color-coded using a red-yellow-green scale, with green indicating better performance.}
	\includegraphics[width=\textwidth]{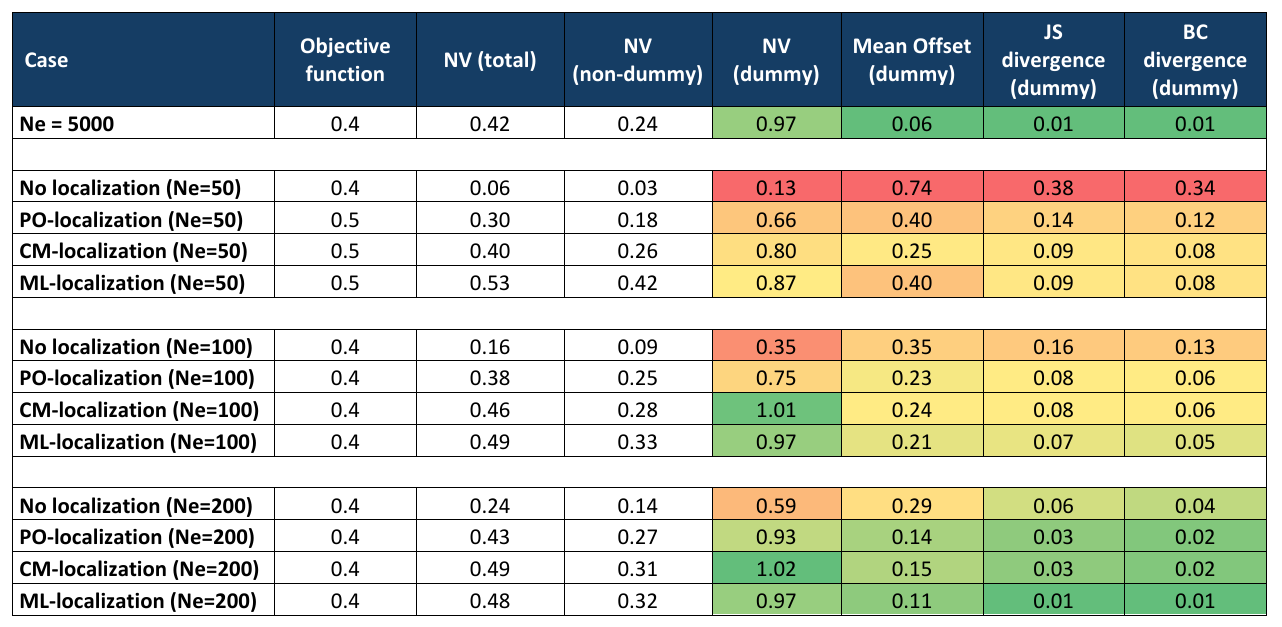}
	\label{tab:test_cases.scalar.data_assim}
\end{table}

\begin{figure}
	\centering
	\includegraphics[width=\textwidth]{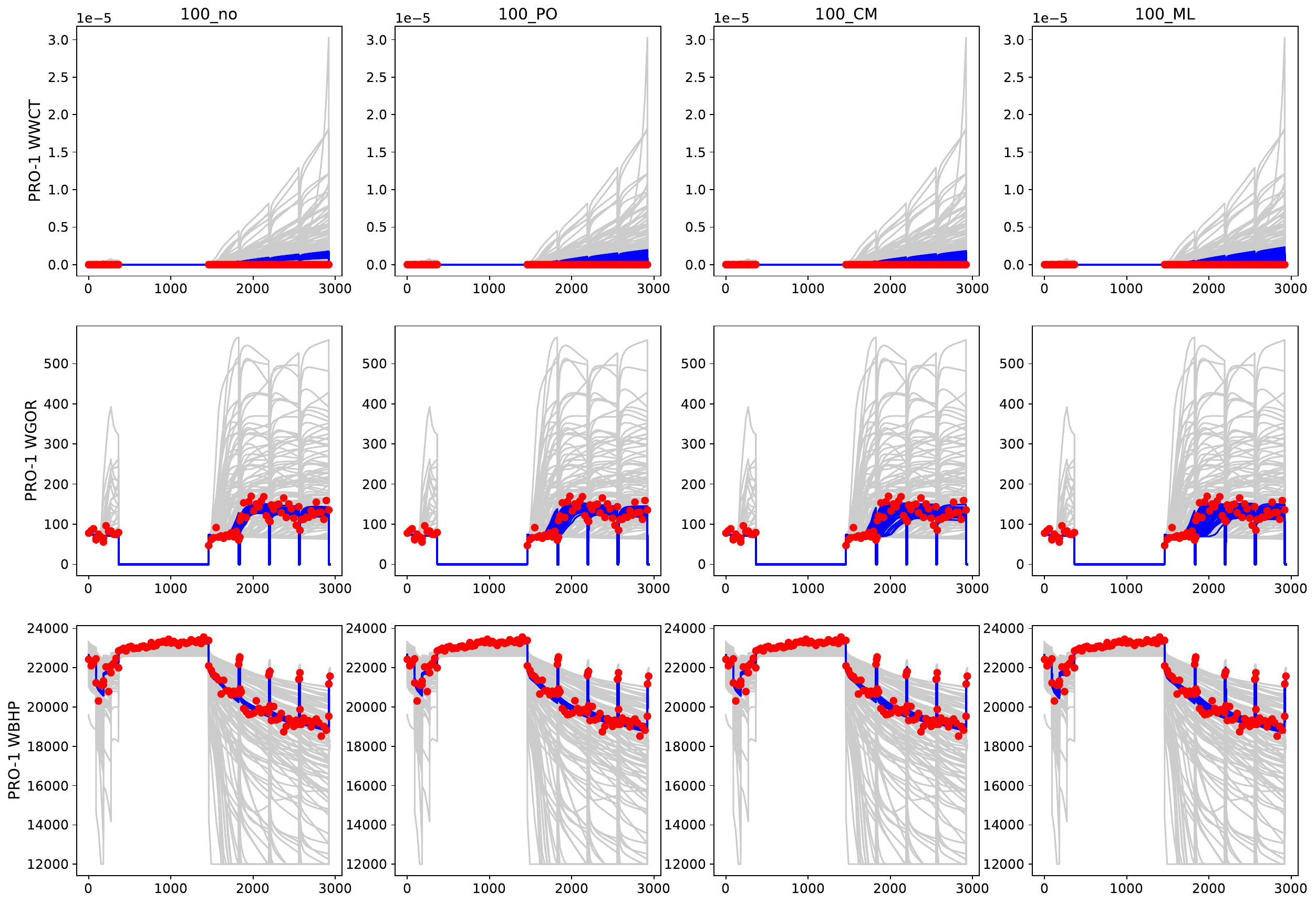}
	\caption{Predicted data for well PRO-1 obtained using the prior ensemble (gray curves) and posterior ensemble (blue curves) for different localization strategies. All results are based on an ensemble size of 100. The first column (100\_no) shows the results without localization, the second column (100\_PO) shows results with PO-localization, the third column (100\_CM) presents results with CM-localization, and the fourth column (100\_ML) shows results with ML-localization. The red dots represent the observed data.}
	\label{fig:test_cases.scalar.data_assim.pro-1}
\end{figure}

The results in Table~\ref{tab:test_cases.scalar.data_assim} show that all cases yielded similar values for the average objective function, which aligns with the expected value for posterior samples in linear-Gaussian problems, which is around 1/2 \citep{oliver:18b}. This indicates that all methods achieved a successful match of the observations. This outcome is further illustrated in Fig.~\ref{fig:test_cases.scalar.data_assim.pro-1}, which presents the observed and predicted data for one of the wells in the field using different localization schemes, with an ensemble size of $N_e = 100$. This figure shows that all cases achieved good agreement between the posterior predictions and the observations.

In terms of NV, the case without localization shows a significant underestimation compared to the reference case with $N_E = 5{,}000$. For $N_e = 50$, only 3\% of the original variance was retained for the non-dummy parameters, indicating near ensemble collapse. The NV for the dummy parameters was also severely underestimated. These results demonstrate that the absence of localization leads to an ensemble that substantially underestimates the posterior uncertainty. PO-localization led to noticeable improvements in NV values, although relevant underestimation persisted for the dummy parameters. In contrast, both CM- and ML-localization yielded systematically higher NV values across all ensemble sizes, for both dummy and non-dummy model parameters. The results for the absolute mean offset, as well as the JS and BC divergences for the dummy variables, indicate that CM- and ML-localization with $N_e = 200$ produced results comparable to those obtained with the much larger ensemble of $N_e = 5{,}000$.

\subsubsection{Sensitivities}
\label{sec:test_cases.scalar.sens}

In this section, we present the results of sensitivity analyses for the proposed localization methods with respect to ML algorithms, ML training, ensemble size, and the combination of CM- and ML-localization. For all sensitivity tests, each data assimilation run is repeated ten times with different prior ensembles.

\begin{figure}
	\centering
	\includegraphics[width=\textwidth]{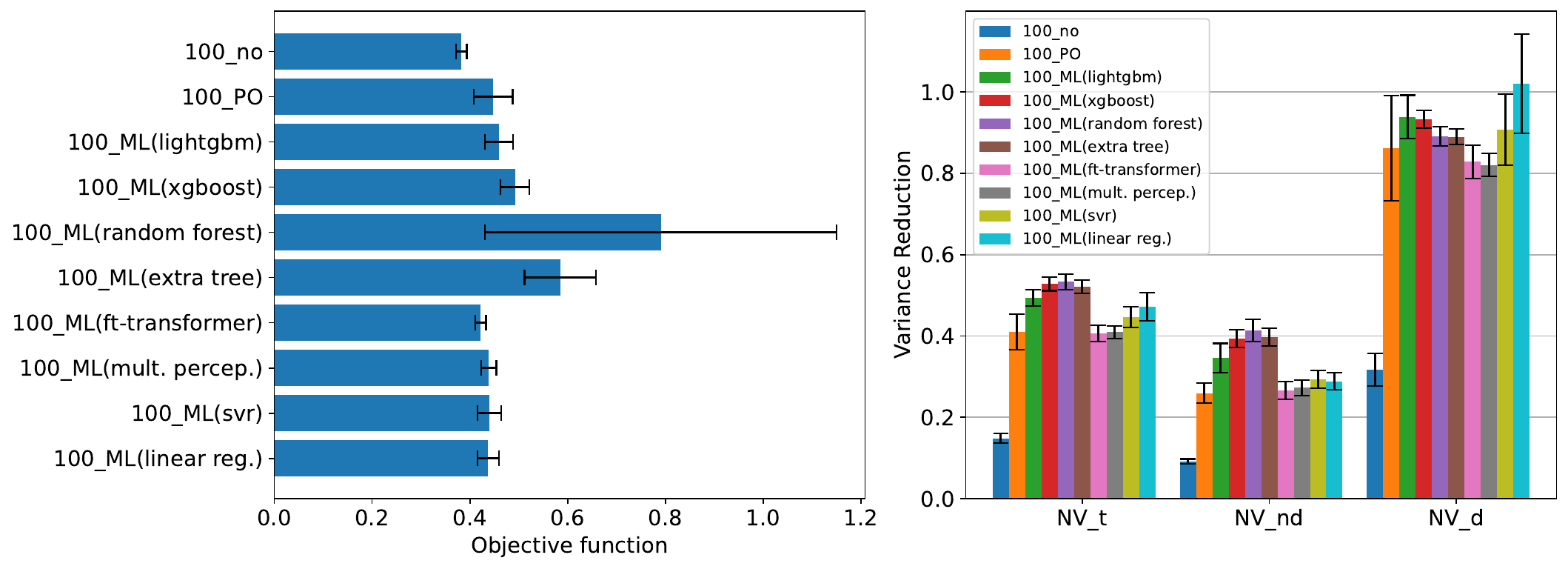}
	\cprotect\caption{Data assimilation results with $N_e=\text{100}$, comparing ML-localization using different machine learning methods. The labels NV\_t, NV\_nd, and NV\_d denote the normalized variance of all parameters, non-dummy parameters, and dummy parameters, respectively.}
	\label{fig:test_cases.scalar.sens.ml_comp}
\end{figure}

Fig.~\ref{fig:test_cases.scalar.sens.ml_comp} presents the data assimilation results obtained using ML-localization with different ML algorithms. For comparison, results without localization and with PO-localization are also included. The plot on the left shows the values of the objective function, while the plot on the right displays the NV for dummy parameters (NV\_d), non-dummy parameters (NV\_nd), and all parameters combined (NV\_t). The results indicate that ML algorithms based on ensemble of decision trees (Random Forest, Extra Trees, XGBoost, and LightGBM), yielded higher NV values for the non-dummy parameters and NV values close to one for the dummy parameters. However, Random Forest and Extra Trees also produced higher data mismatch objective values, both in terms of mean and standard deviation, suggesting reduced robustness compared to the other two methods. 

The superior performance of ensemble of decision tree models is consistent with the findings of \citet{grinsztajn:22a}, which show that these models outperform others when the data exhibit certain characteristics: the presence of uninformative features, non-smooth target functions, and features that are not invariant under rotation. In our case, and typically in most reservoir applications, uninformative features are present, as not all model parameters influence all observed data points. Additionally, the dataset is not rotationally invariant, meaning that linear combinations of features do not preserve the relevant information.
Among all ML methods tested in this study, those based on gradient boosting decision trees (XGBoost and LightGBM) showed the best overall performance. We chose to continue the experiments in this section using LightGBM. 
 
\begin{figure}
	\centering
	\includegraphics[width=\textwidth]{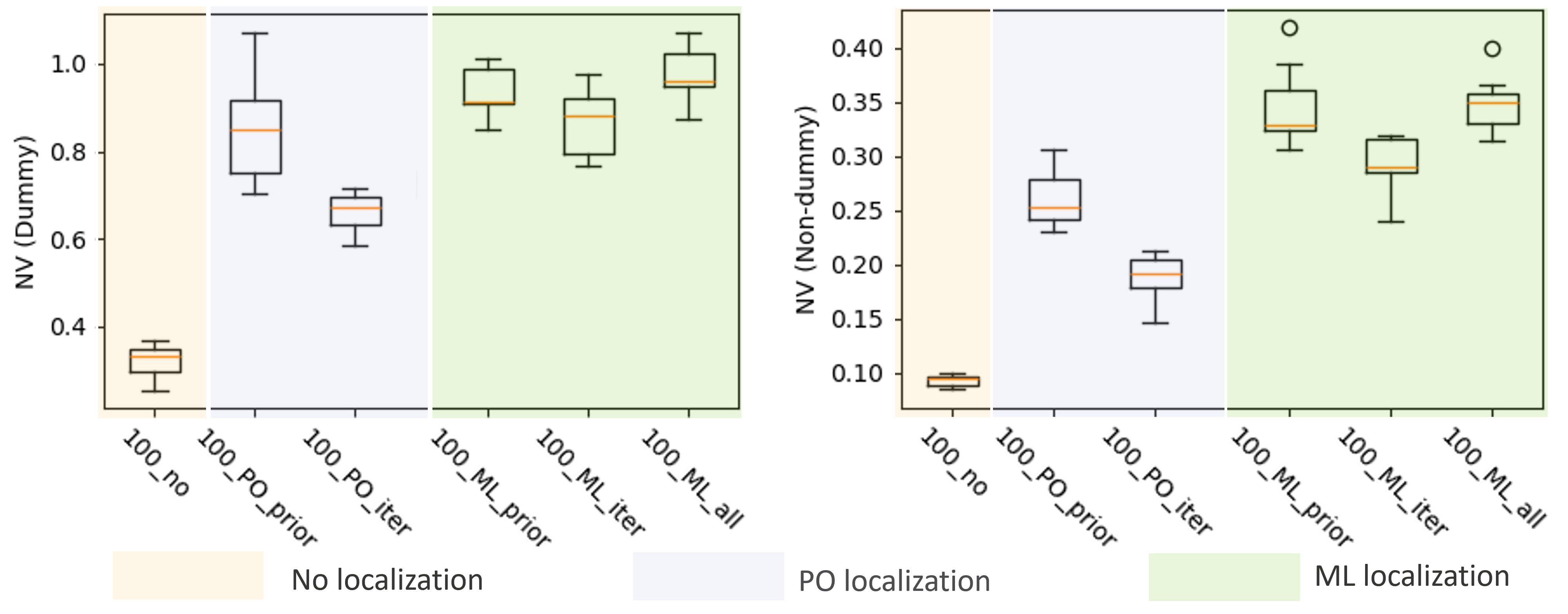}
	\caption{Data assimilation results with $N_e =\text{100}$, comparing the NV for different strategies used to compute localization coefficients. The label 100\_no denotes the case without localization. The labels 100\_PO-prior and 100\_PO-iter refer to PO-localization using the prior ensemble and updating the localization coefficients at each ES-MDA iteration, respectively. The labels 100\_ML-prior, 100\_ML-iter, and 100\_ML-all correspond to ML-localization using a proxy trained on the prior ensemble, the current iteration ensemble, and all previous ensembles, respectively.}
	\label{fig:test_cases.scalar.sens.prior-iter}
\end{figure}

Fig.~\ref{fig:test_cases.scalar.sens.prior-iter} compares ML-localization (using LightGBM) with PO-localization and no localization, considering different strategies for computing the localization coefficients. In the first strategy, labeled ``prior,'' the localization coefficients are computed using only the prior ensemble, this is the standard approach proposed in Algorithm~\ref{algo:method.ml-local} and used in the previous results. In the second strategy, labeled ``iter,'' the coefficients are updated at each ES-MDA iteration.
Conceptually, the ``iter'' strategy should be preferred, as it updates the localization coefficients based on the current ensemble, potentially capturing changes during the assimilation process. However, \citet{lacerda:19a} reported better results when keeping the localization coefficients fixed using only the prior ensemble, possibly due to the larger variance present in the prior. In our tests, the results for PO-localization confirm these findings, as the ``prior'' strategy yields higher NV values for both dummy and non-dummy parameters. Although not shown here, it is worth noting that all data assimilation runs presented in this figure achieved similar values for the data mismatch objective function.

For ML-localization, the localization coefficients are always computed using the large ensemble with $N_E = 5{,}000$ members, which is not updated. In this context, the ``iter'' strategy refers to retraining the ML proxy at each ES-MDA iteration using the current ensemble as the training dataset. We also consider a third strategy, labeled ``all,'' in which the ML proxy is retrained using the accumulated data from all previous iterations. The rationale is that enlarging the training set may improve the quality of the ML model.

The results in Fig.~\ref{fig:test_cases.scalar.sens.prior-iter} show that the ``prior'' strategy outperforms the ``iter'' strategy and achieves results comparable to the ``all'' strategy, but without the additional computational cost associated with repeated training.

\begin{figure}
	\centering
	\includegraphics[width=\textwidth]{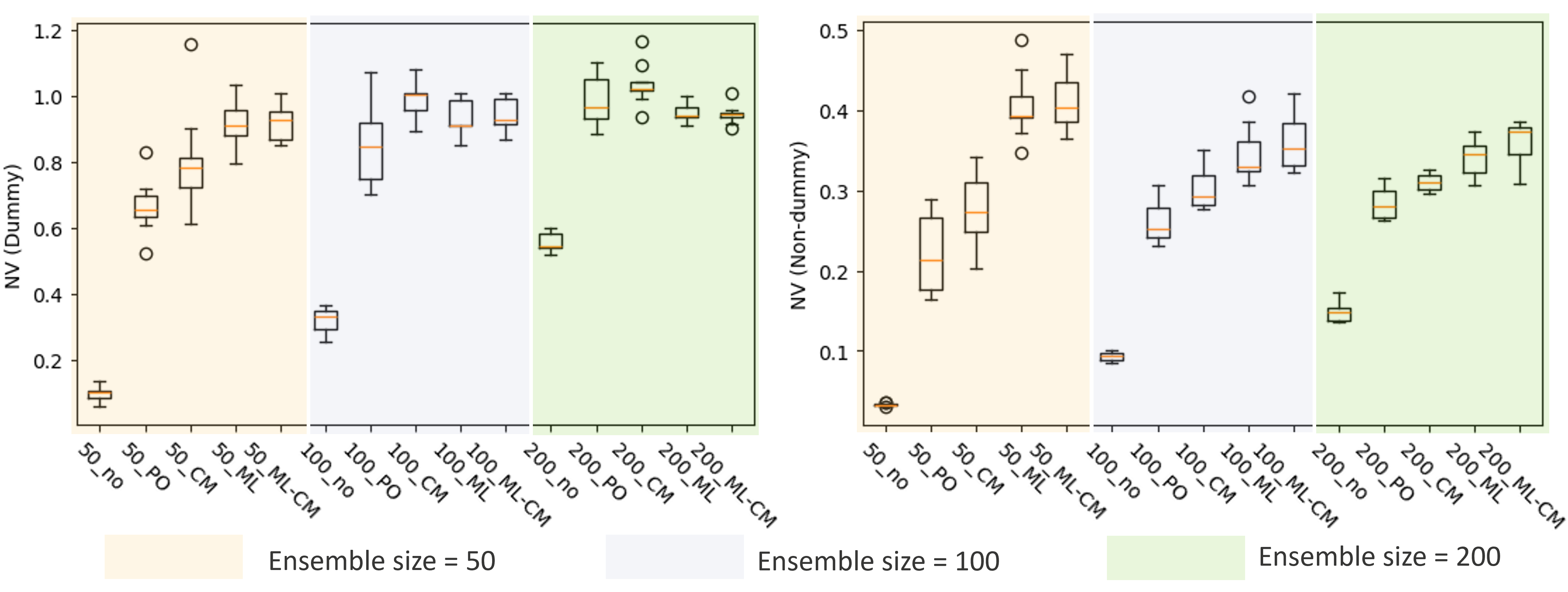}
	\caption{Data assimilation results comparing the localization schemes with different ensemble sizes. The label 50\_no denotes the case without localization with $N_e=\text{50}$. The label 50\_PO refer to PO-localization, 50\_ML to ML-localization, 50\_CM to CM-localization, and 50\_ML-CM to applying CM- and ML-localization combined.}
	\label{fig:test_cases.scalar.sens.ens_size}
\end{figure}

Fig.~\ref{fig:test_cases.scalar.sens.ens_size} compares the localization methods in terms of NV for different ensemble sizes. As expected, the NV increases with ensemble size across all cases, reflecting the reduction in spurious correlations with larger ensembles. Both CM- and ML-localization yield high NV values. Interestingly, for dummy parameters, CM-localization produces NV values closer to the reference value of one, whereas for non-dummy parameters, ML-localization performs better. These results suggest that a combination of CM- and ML-localization may be beneficial. We added this combined strategy in Fig.~\ref{fig:test_cases.scalar.sens.ens_size} and the results show a slight improvement in NV for the non-dummy parameters. As in previous cases, all strategies exhibit similar performance in terms of the data mismatch objective function.

\subsection{Grid Parameters}
\label{sec:test_cases.grid}

In this section, we evaluate the performance of ML-localization in problems where the model parameters are defined on a grid. This is the typical setting in reservoir history matching, when the goal is to update realizations of rock properties such as porosity and permeability. In such cases, distance-based localization is the standard approach \citep[Chap.~7]{emerick:25bk}, and is therefore included in our comparisons. Note that for grid-based parameters, prior realizations are generated using geostatistical workflows, and the prior covariance matrix is never explicitly constructed. In fact, for operational models with a large number of gridblocks, storing this matrix would be impractical. As a result, CM-localization is not considered in this section.

We consider two test problems. The first is a small two-dimensional reservoir model representing a CO$_\text{2}$ storage scenario. The second is the PUNQ-S3 model, where porosity, horizontal permeability, and vertical permeability at the gridblock level are treated as model parameters.

\subsubsection{Grid Case 1: CCS Model}
\label{sec:test_cases.grid.ccs}

The test case corresponds to a small reservoir model undergoing CO$_\text{2}$ injection, representing a CCS scenario. This is a similar problem used in \citet{seabra:24a}. Simulations were performed using the open-source Delft Advanced Research Terra Simulator (open-DARTS) \citep{voskov:24a}. The reservoir model uses a structured Cartesian grid with dimensions of 32~$\times$32$~\times$~1 cells, representing a two-dimensional horizontal layer. Each gridblock is uniformly sized at 192 meters in length and width and 10 meters in thickness. The simulated layer is located at a depth of 2,000 meters, consistent with typical geological settings for CO$_\text{2}$ storage in a saline aquifer, and is initially at a pressure of 200~bar.

\begin{figure}
\centering
  \includegraphics[width=0.5\textwidth]{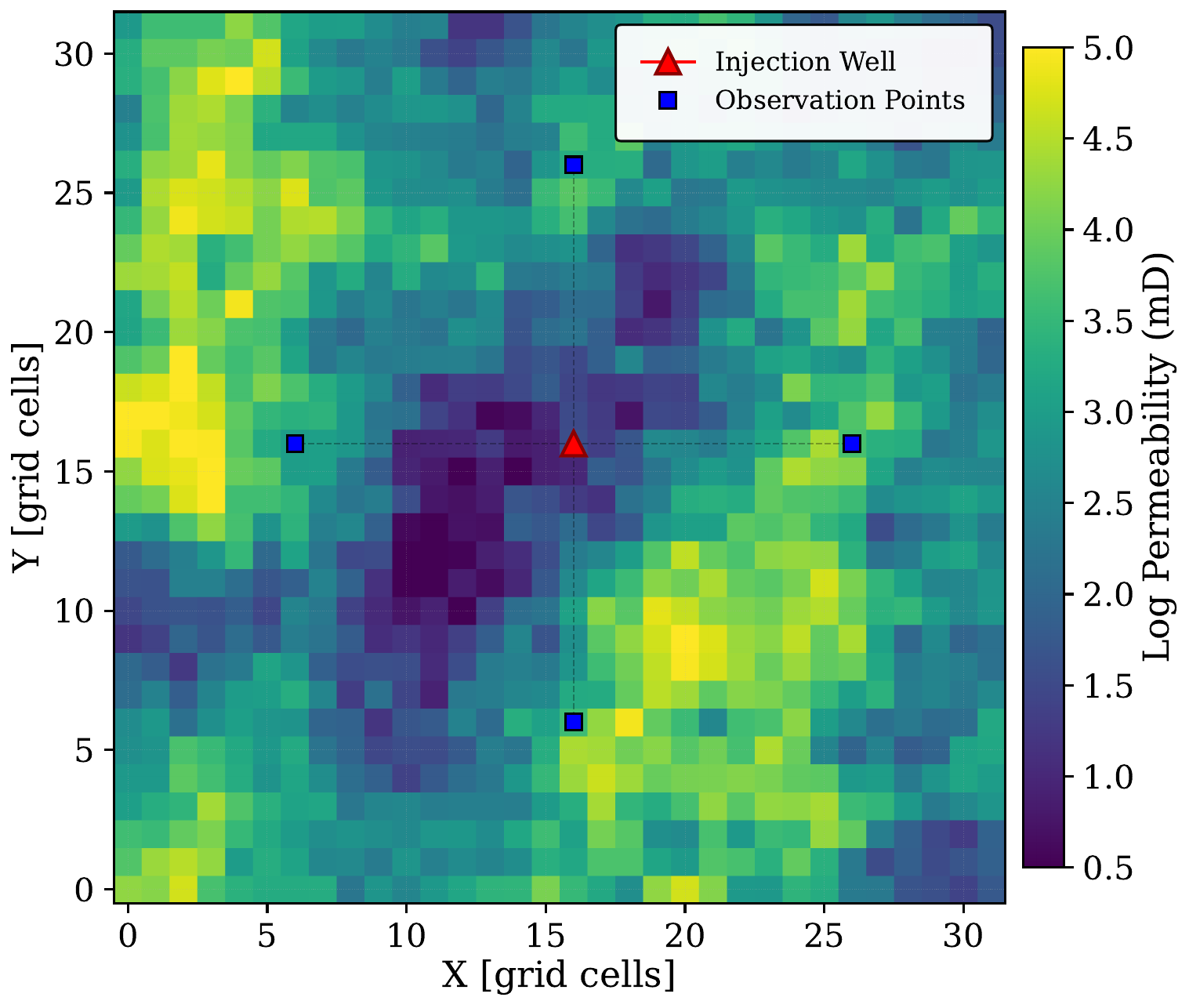}
\caption{Reference permeability field for the CCS model. The red star indicates the CO$_\text{2}$ injection well, and the blue squares denote the positions of the four pressure monitoring wells.}
\label{fig:test_cases.grid.ccs.true}
\end{figure}

The reservoir simulation considered a two-component, two-phase system comprising CO$_\text{2}$ and water. The prior log-permeability fields were generated from a Gaussian distribution with a mean of 3.0 and a variance of 1.0. The spatial correlation was modeled using a Gaspari-Cohn function with a correlation range corresponding to the size of 10 grid cells. Realizations were generated from the resulting covariance matrix via Cholesky decomposition. Porosity was assumed to be constant throughout the reservoir at 25\%. Fig.~\ref{fig:test_cases.grid.ccs.true} shows the reference (ground truth) permeability field, highlighting the location of a single CO$_\text{2}$ injection well at the center of the model and four pressure monitoring wells.

The data assimilation experiments used ensembles of realizations of log-permeability (a total of 1,024 model parameters), generated using the same prior settings as those used to construct the ground truth. The observations consisted of pressure measurements at the monitoring well locations over a two-year period, resulting in a total of 96 data points. We perform the data assimilation using the ES-MDA with four iterations with no localization, distance-based localization (DB-localization), PO-localization, and ML-localization. For the distance-based localization case, we used the standard Gaspari-Cohn function with a critical length corresponding to the size of 10 gridblocks.

\begin{figure}
	\centering
	\includegraphics[width=0.9\textwidth]{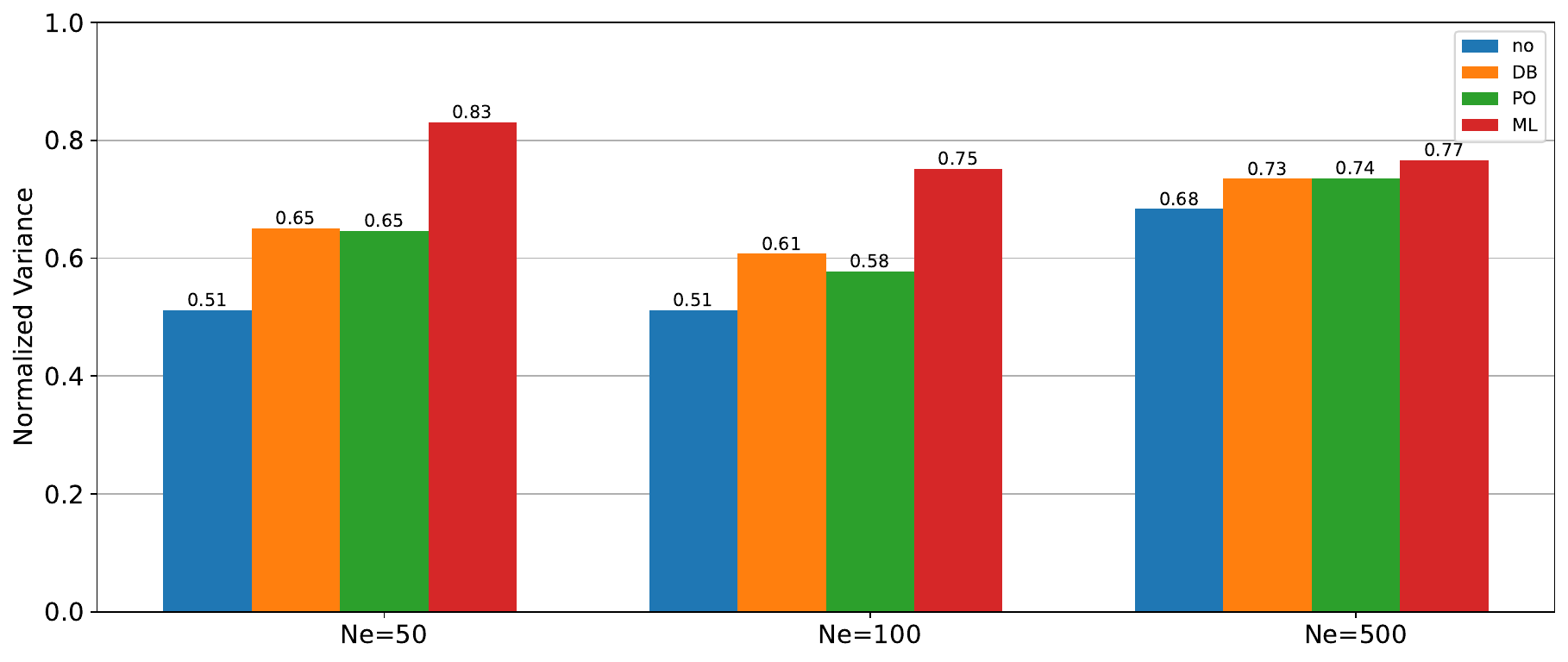}
	\caption{Normalized variance results with varying ensemble sizes for the CCS Model. This figure includes the results of data assimilation without localization (no), DB-localization (DB), PO-localization (PO), and ML-localization (ML).}
	\label{fig:test_cases.grid.ccs.nv}
\end{figure}

\begin{figure}
	\centering
	\includegraphics[width=0.9\textwidth]{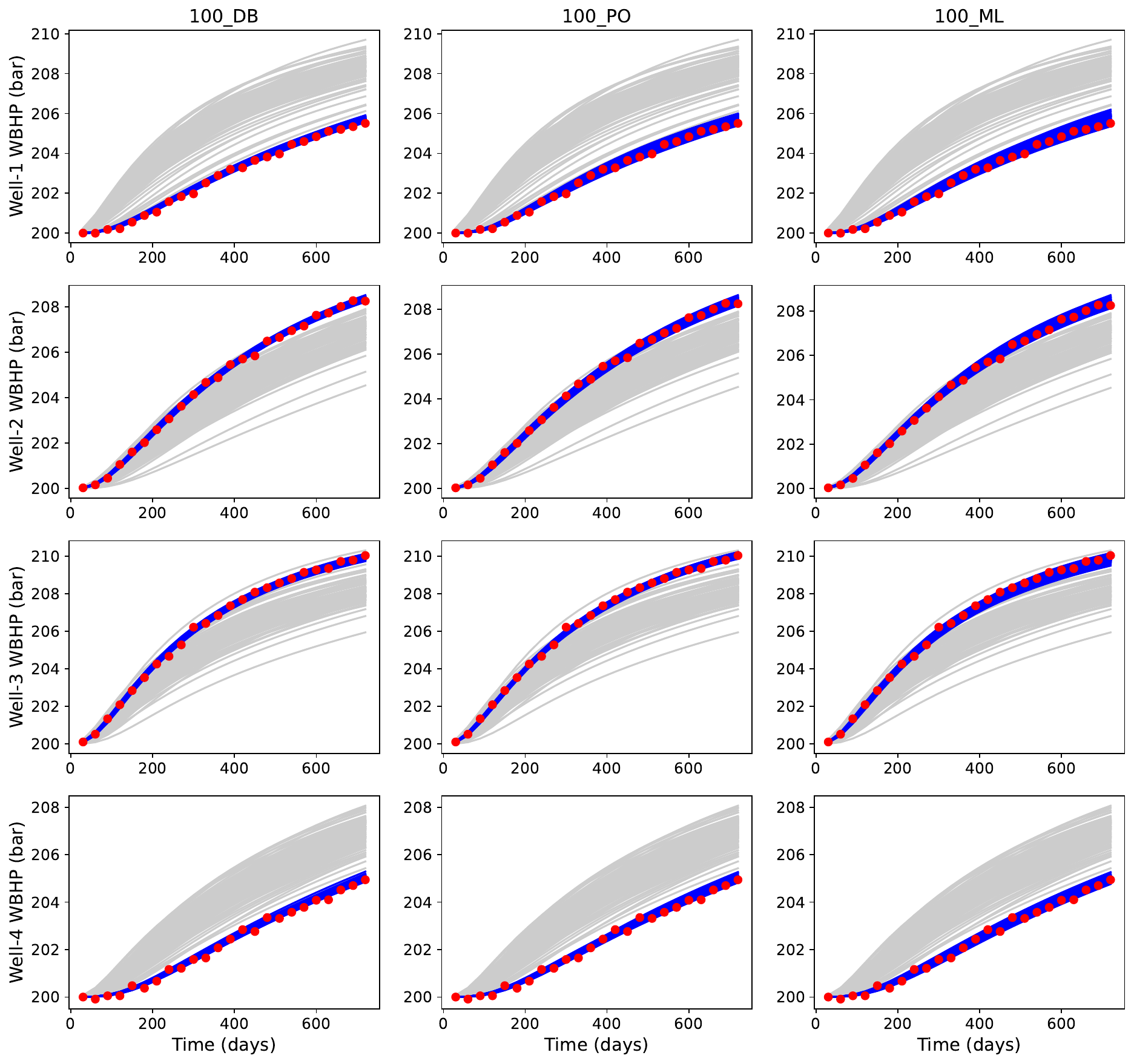}
	\caption{Predicted WBHP data before and after data assimilation for an ensemble size of $N_e=\text{100}$. Each row represents a monitoring well and each column a different localization scheme. The red dots represent the observed data, the gray curves the prior ensemble, and the blue curves the posterior ensemble.}
	\label{fig:test_cases.grid.ccs.bhp}
\end{figure}

Fig.~\ref{fig:test_cases.grid.ccs.nv} shows the values of NV obtained after data assimilation with different ensemble sizes. For ML-localization we used the XGBoost method. The results in this figure indicate that ML-localization resulted in larger NV for all ensemble sizes. In terms of data match quality, all methods resulted in similar values of data mismatch. This is illustrated in Fig.~\ref{fig:test_cases.grid.ccs.bhp}, which shows the predicted well bottom-hole pressure (WBHP) for all localization cases with an ensemble of 100 realizations.

\begin{figure}
	\centering
	\includegraphics[width=\textwidth]{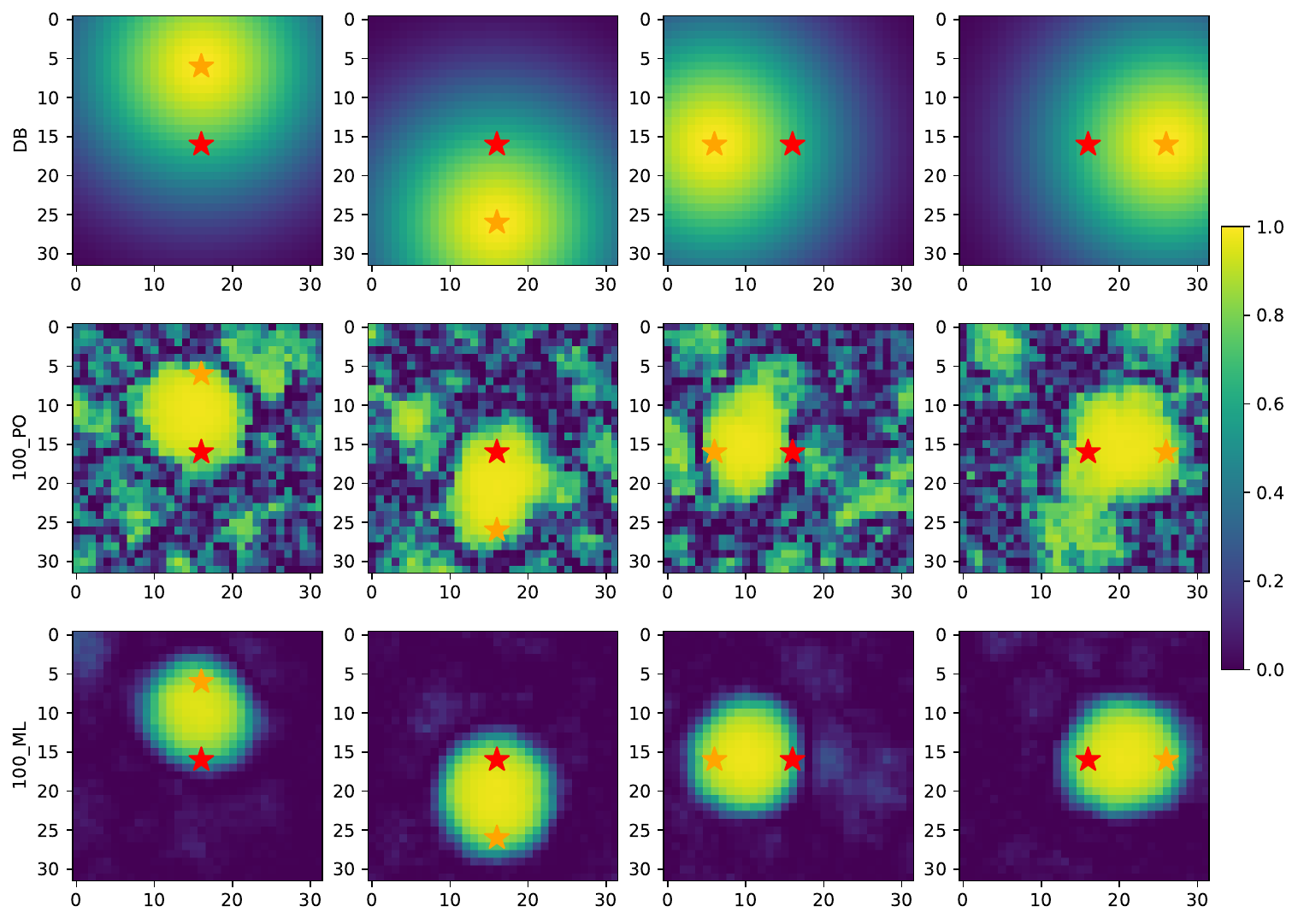}
	\caption{Localization values ($N_e = \text{100}$) for the CCS model. Each column corresponds to a different monitoring well, and each row shows results from a different localization scheme. The red star indicates the CO$_\text{2}$ injection well, and the orange stars denote the positions of the monitoring wells.}
	\label{fig:test_cases.grid.ccs.local}
\end{figure}

Fig.~\ref{fig:test_cases.grid.ccs.local} shows the localization maps for each monitoring well at the final time step. Each image includes the location of the CO$_\text{2}$ injection well (red star) and the corresponding monitoring well (orange star). Distance-based localization produces a smooth spatial distribution, with values decaying from one at the data location to zero in regions far from the well. PO-localization produces patterns where the regions between injection and monitoring wells are clearly emphasized. However, these maps also show high variability in areas far from the wells, likely due to spurious correlations. In contrast, ML-localization effectively suppresses distant correlations, yielding clean localization maps concentrated along the region connecting each pair of wells. Notably, DB-Localization is inherently centered on the data location, whereas PO- and ML-localization do not follow this constraint. Additionally, ML-localization exhibits a much sharper decay in localization values.

\clearpage

\subsubsection{Grid Case 2: PUNQ-S3 Model}
\label{sec:test_cases.grid.punq}

Here, we consider the PUNQ-S3 model with grid parameters corresponding to the spatial distribution of porosity, horizontal and vertical permeability, resulting in 5,283 grid parameters updated during data assimilation. The prior ensemble was generated using sequential Gaussian simulation \citep{deutsch:02bk} with settings following the model description presented in \citep{floris:01}. The observed data correspond to the same data described in Section~\ref{sec:test_cases.scalar}. The data assimilation experiments considered an ensemble of 100 realizations with four ES-MDA iterations.

\begin{figure}[h]
	\centering
	\includegraphics[width=\textwidth]{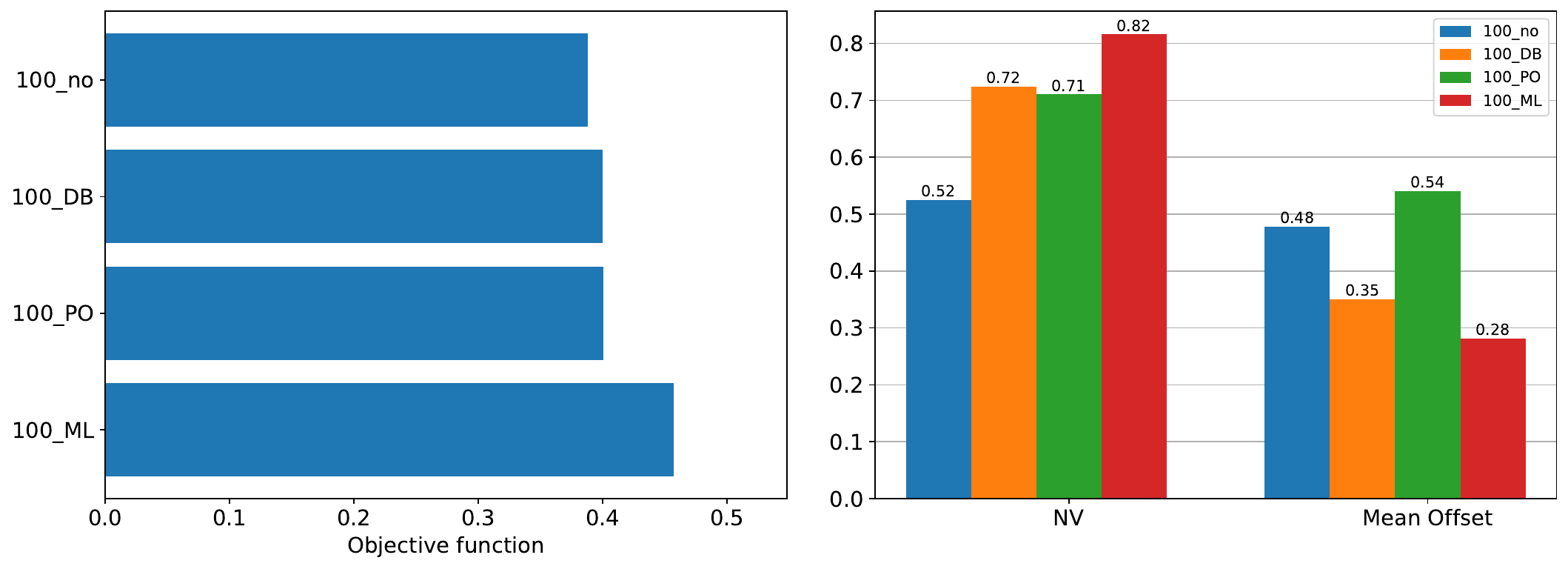}
	\caption{Data assimilation results for the PUNQ-S3 with grid model parameters ($N_e=\text{100}$).}
	\label{fig:test_cases.grid.punq.of-nv}
\end{figure}

Fig.~\ref{fig:test_cases.grid.punq.of-nv} presents the results in terms of the data mismatch objective function, NV, and mean offset after data assimilation using no localization, DB-localization, PO-localization, and ML-localization. For the DB-localization, we used the Gaspari-Cohn function with critical length equal to the size of 10 gridblocks. Alternative correlation lengths were tested, and this value provided the best trade-off, yielding high NV with an acceptable average objective function (below 0.5). For ML-localization we used LightGBM. The results show that ML-localization achieved the best overall performance, producing an ensemble with higher NV and lower mean offset, while maintaining the mean objective function below 0.5.

\begin{figure}
    \centering
    \begin{subfigure}{\textwidth}
        \includegraphics[width=\textwidth]{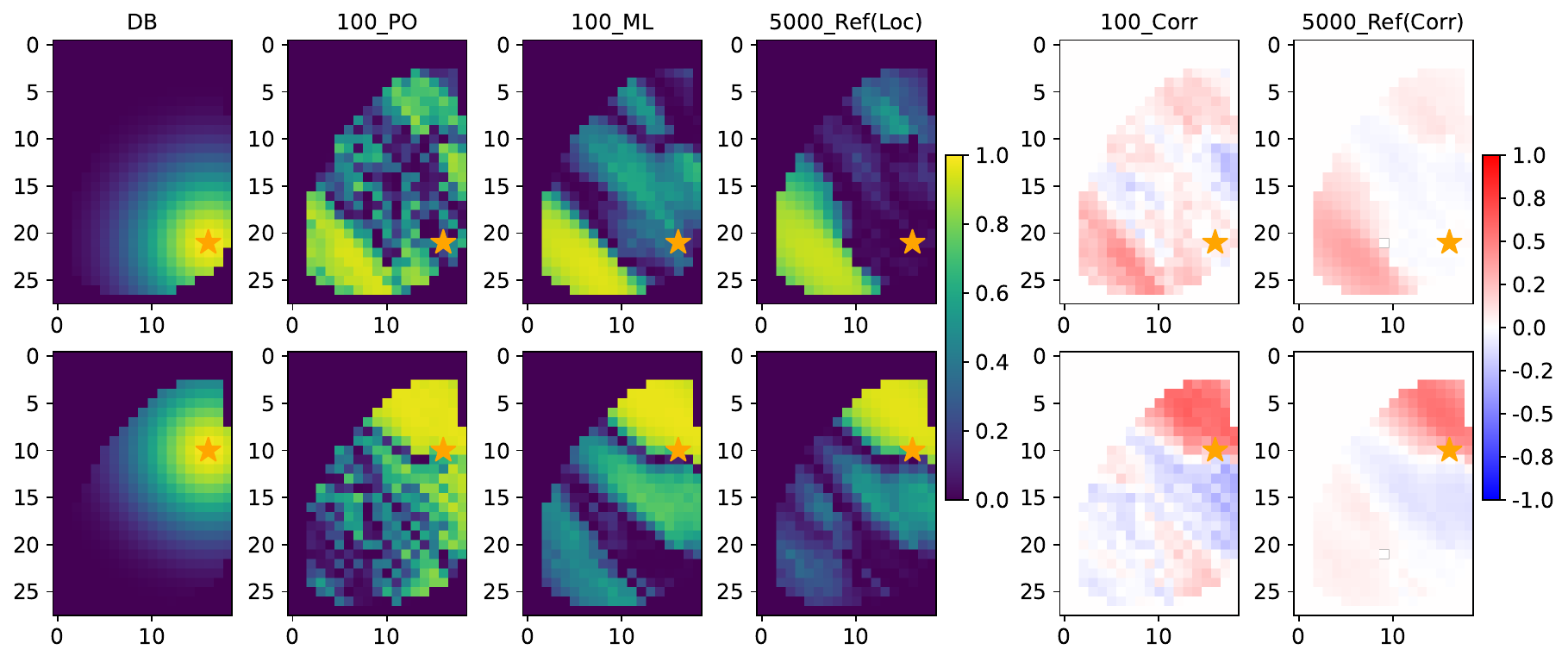}
        \caption{WBHP}
    \end{subfigure}
    \hfill
    \begin{subfigure}{\textwidth}
        \includegraphics[width=\textwidth]{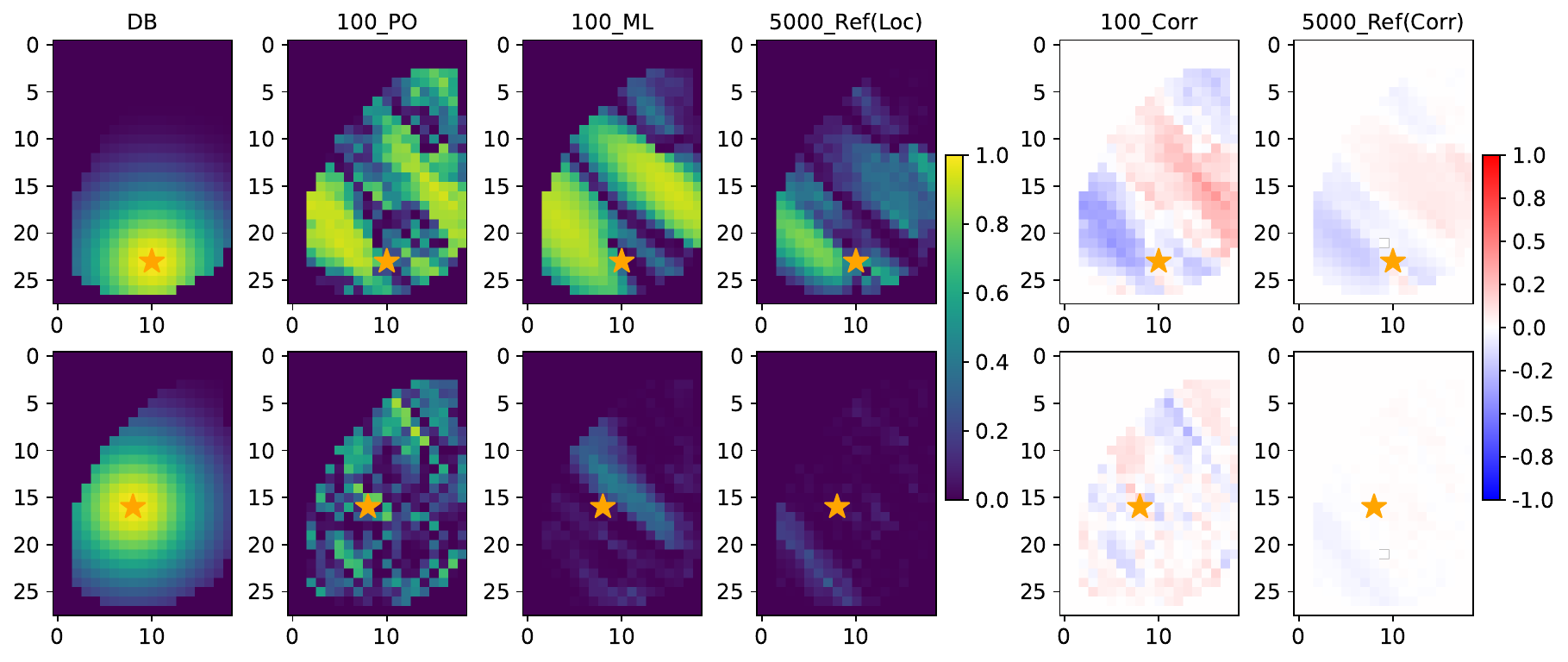}
        \caption{WWCT}
    \end{subfigure}
    \hfill
    \begin{subfigure}{\textwidth}
        \includegraphics[width=\textwidth]{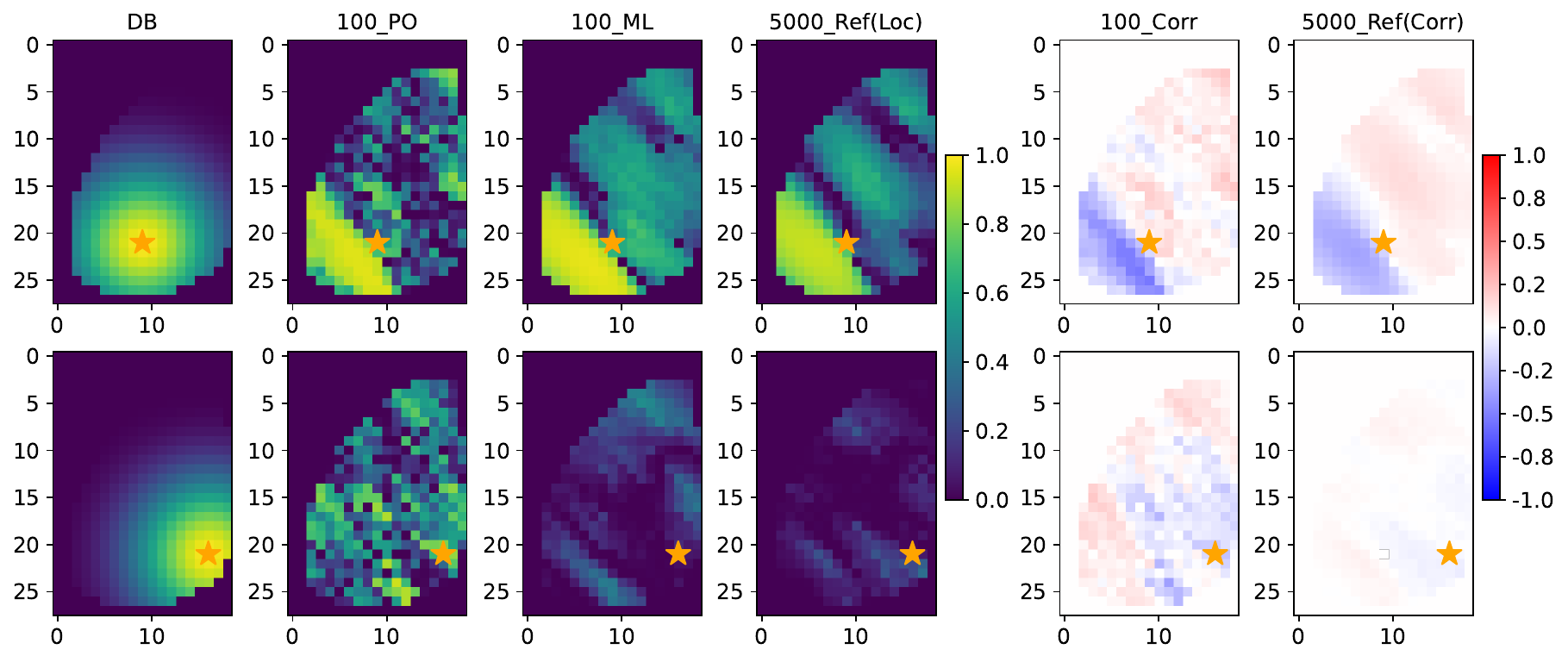}
        \caption{WGOR}
    \end{subfigure}
    \caption{Localization values for horizontal log-permeability in the PUNQ-S3 case with grid-based parameters. Each row corresponds to a different well, indicated by the orange star. In each subfigure: DB denotes DB-localization, 100\_PO indicates PO-localization, 100\_ML indicates ML-localization, and 5000\_Ref(loc) refers to localization computed using the reference ensemble with 5,000 realizations. 100\_Corr and 5000\_Ref(Corr) represent the correlation coefficients computed using ensembles of 100 and 5,000 members, respectively.}
    \label{fig:test_cases.grid.punq.local}
\end{figure}

Fig.~\ref{fig:test_cases.grid.punq.local} shows localization maps for the first layer of the model, focusing on horizontal permeability and considering data points from WBHP (Fig.~\ref{fig:test_cases.grid.punq.local}a), WWCT (Fig.~\ref{fig:test_cases.grid.punq.local}b), and WGOR (Fig.~\ref{fig:test_cases.grid.punq.local}c). In all cases, the selected data correspond to the last time step of each time series. For WBHP, this point represents a static pressure measurement during a shut-in period. For comparison, we also include localization maps computed using the PO taper function, where cross-covariances were estimated from $N_E = 5{,}000$ reservoir simulations. This reference case is labeled as 5000\_Ref(loc) in Fig.~\ref{fig:test_cases.grid.punq.local}. It represents the performance limit expected for ML-localization, as it uses the same 5,000 realizations to compute cross-covariances, but based on actual reservoir simulations rather than proxy predictions. Each plot also includes correlation coefficient maps estimated with both 100 and 5,000 reservoir simulations.

The results in Fig.~\ref{fig:test_cases.grid.punq.local} show that ML-localization is capable of producing localization values similar to those obtained from the reference case with 5,000 simulations. Moreover, the resulting localization patterns are far more complex than the circular or elliptical regions centered at the well location typically assumed in practical applications of DB-localization \citep[Chap.~7]{emerick:25bk}.

First, we observe that the localization regions are strongly influenced by the prior distribution of log-permeability, which exhibits a major correlation direction oriented at -45$^\circ$. These findings are consistent with the discussion in \citep{emerick:11e}, where it is suggested that the localization function should reflect a combination of prior correlations and data sensitivity.

Second, the ML-localization maps are not necessarily centered at the data locations. In some cases, the well with data lies outside the region with the highest localization values. Although this may appear counterintuitive, it is a consequence of the prior realizations being conditioned on porosity and permeability measurements at the well locations. This leads to significantly reduced prior variance around the wells, which in turn affects the localization values. However, the fact that DB-localization yields large localization coefficients around these wells does not pose a significant issue. Since the prior variance in these regions is low, the corresponding updates remain small even with high localization values. The main limitation of DB-localization in this case is that it produces small (or even zero) localization coefficients at locations far from the well, where relevant cross-correlations are actually present, as indicated by the results labeled 5000\_Ref(Corr).

Finally, it is noteworthy that in the second rows of Figs.~\ref{fig:test_cases.grid.punq.local}b and \ref{fig:test_cases.grid.punq.local}c, the ML-localization values are close to zero across the entire model. This is because the predicted data from the prior ensemble exhibit very low variance—these wells show no water or gas breakthrough in either the observed dataset or the prior predictions. In contrast, the localization maps obtained using PO-localization display several spurious nonzero values.

\begin{figure}
	\centering
	\includegraphics[width=\textwidth]{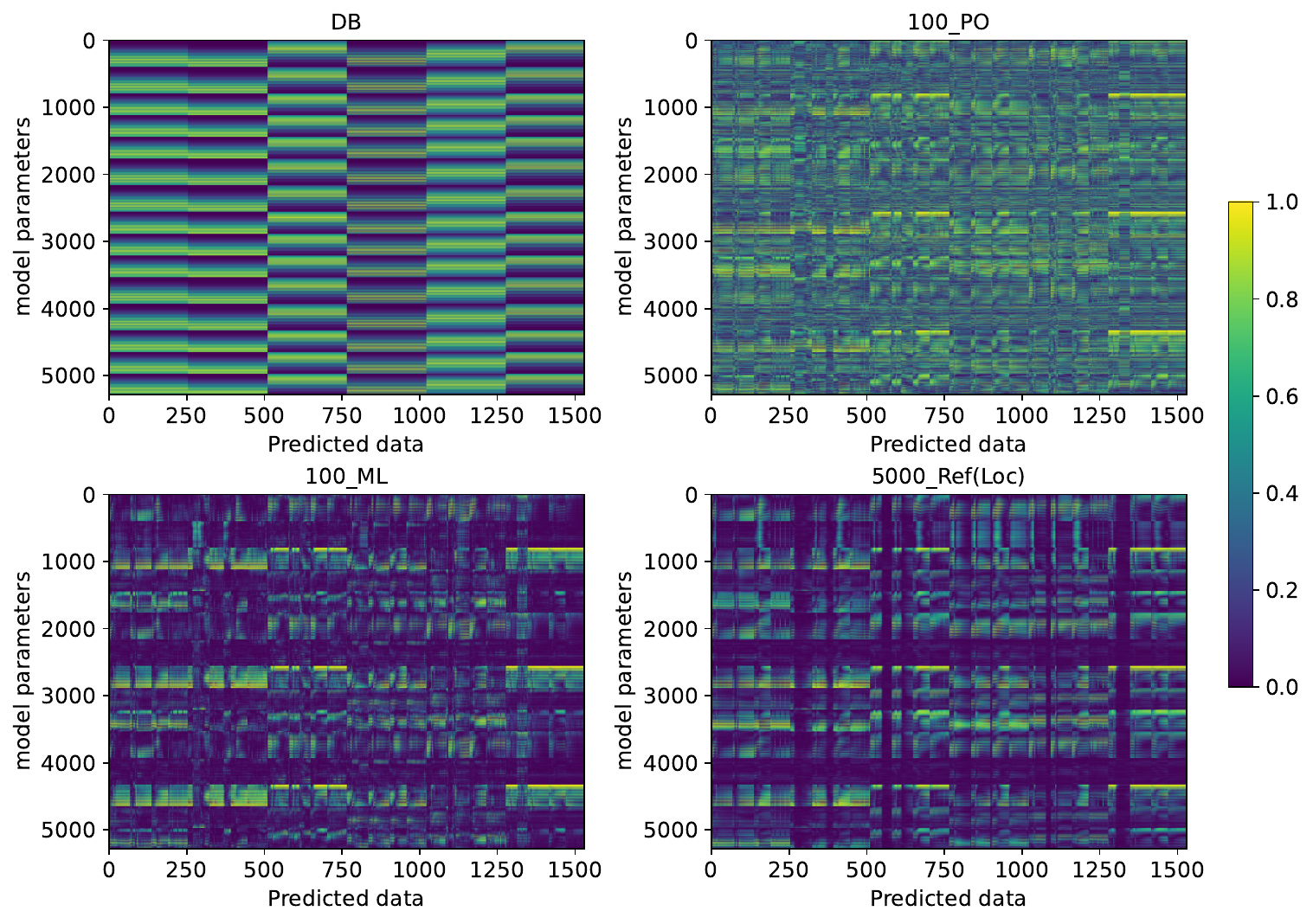}
	\caption{Localization matrices calculated using the distance-based (DB) localization, PO-localization, and ML-localization with an ensemble of $N_e=\text{100}$. A reference localization generated using 5,000 runs is also shown.}
	\label{fig:test_cases.grid.punq.local-matrix}
\end{figure}

Fig.~\ref{fig:test_cases.grid.punq.local-matrix} presents the full localization matrices computed using DB-localization, PO-localization, ML-localization, and the reference localization generated from 5,000 simulations. The reference matrix reveals that many pairs of model parameters and predicted data should have zero correlation. PO-localization retains several spurious correlations, while DB-localization suppresses correlations in some regions that may contain useful information and retains correlations in regions where they are not warranted. In contrast, ML-localization produces a matrix that closely resembles the reference localization (5000\_Ref(Loc)), effectively preserving meaningful correlations while removing spurious ones.
 
\clearpage

\subsection{Computational Cost of the ML Proxy}
\label{sec:test_cases.comp}

In this section, we investigate the computational cost associated with training the ML proxies. We concentrate in the gradient boosting methods (XGBoost and LightGBM) since they produced the best overall performance. Although we do not perform any additional reservoir simulation rather than the ones required by the data assimilation process, when applying the ML-localization there are the additional costs of training the ML and predicting the data points of the large ensemble. 

\begin{figure}[h]
	\centering
	\includegraphics[width=0.9\textwidth]{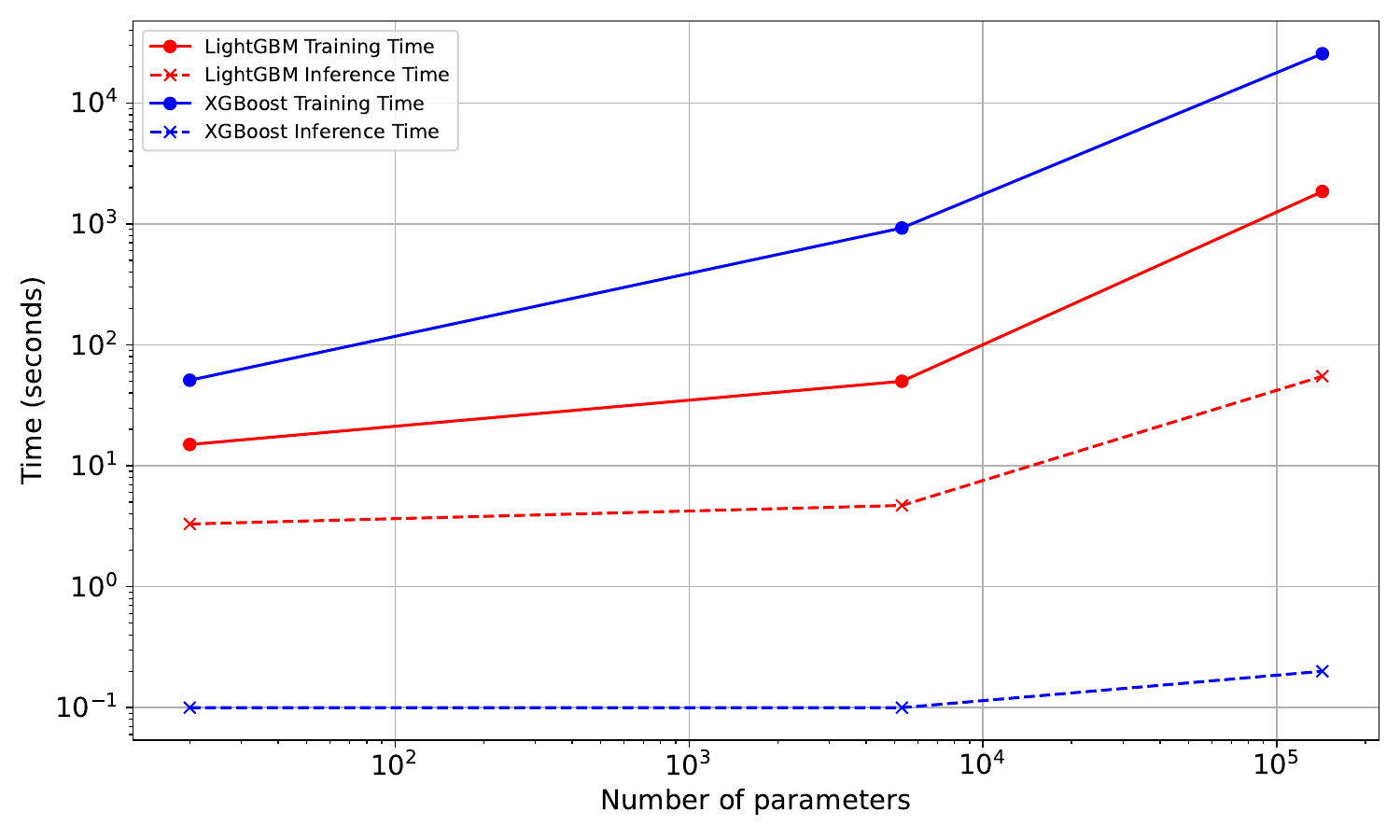}
	\caption{Training and inference/prediction time of the LightGBM and XGBoost for different number of model parameters. All cases consider $N_e=\text{100}$, $N_e=\text{5,000}$, and $N_d=\text{1,530}$.}
	\label{fig:test_cases.comp.cpu}
\end{figure}

Fig.~\ref{fig:test_cases.comp.cpu} shows the training and inference/prediction times for the ML methods as a function of the number of model parameters in the PUNQ-S3 model. We consider three versions of this problem: the first two correspond to the cases used in Sections~\ref{sec:test_cases.scalar} and \ref{sec:test_cases.grid.punq}, with 20 and 5,283 parameters, respectively. The third case is a refined version of the PUNQ-S3 model, containing a total of 142,641 model parameters. All cases use an ensemble size of $N_e = 100$ and a fixed number of data points, $N_d = 1{,}530$. We kept the number of data points constant because, in most practical cases, the number of model parameters can vary by orders of magnitude depending on the parameterization and the size of the subsurface reservoir, while the number of production data points typically remains on the order of thousands. It is important to note that ML training is performed independently for each data point, meaning that the overall training time scales linearly with the number of data points. All tests were performed in a machine with 64 cores of Intel(R) Xeon(R) Platinum 8358 CPU @ 2.60GHz with 256GB of RAM.  

Comparing training times, LightGBM is nearly an order of magnitude faster than XGBoost. However, XGBoost outperforms LightGBM in inference time. Since the total computational cost is the sum of training and inference times, and Fig.~\ref{fig:test_cases.comp.cpu} shows that inference time is only a small fraction of training time, the total cost is primarily determined by the training phase. As a result, LightGBM yields the lowest overall computational cost compared to XGBoost. It is worth noting that all training and inference in this study were performed using a CPU. If a GPU were used, both training and prediction times could be significantly reduced.

\section{Discussion}
\label{sec:disc}

We proposed a novel framework that leverages machine learning to compute distance-free localization coefficients for ensemble-based data assimilation methods. The approach is general and easy to implement, as it employs standard, publicly available machine learning libraries. Importantly, it requires no additional reservoir simulations beyond those already performed during the data assimilation process and it also requires no hyperparameter tuning.

Inspired by the use of linear regression as a proxy, we also proposed a simple correction to the prior cross-covariance used in computing the localization coefficients. Our results show that this procedure leads to significant improvements in data assimilation performance for problems with a small number of parameters. In such cases, this correction can be effectively combined with the proposed machine learning method.

We tested the proposed methods using reservoir data assimilation problems with both scalar and grid-based model parameters. For the scalar case, we evaluated the performance of various machine learning methods in improving cross-covariance estimates and data assimilation outcomes. The results indicates that gradient boosting decision tree methods, such as XGBoost and LightGBM, provided a favorable balance between training time and covariance accuracy. Moreover, the machine learning-based localization yielded higher posterior ensemble variance compared to both the direct use of the pseudo-optimal taper function and the case without localization, while maintaining comparable data match quality.

For the grid-based parameter cases, the machine learning-based localization outperformed both the standard distance-based and pseudo-optimal localization methods in terms of data assimilation results. The experiments also showed that the machine learning proxy was able to reproduce localization values similar to those obtained using a large reference ensemble of 5,000 realizations.

A potential limitation of the machine learning-based localization approach is the increasing training cost as the number of model parameters grows. Our experiments showed that the largest case, with 142,641 model parameters, required approximately 30 minutes of training with LightGBM on a CPU using 64 cores. Although this training is performed only once during the data assimilation process, the training time may still become a limiting factor in practice, particularly for problems involving $O(10^6\text{--}10^7)$ model parameters. One potential solution is to leverage GPUs for training, as the ML libraries used are fully GPU-compatible. Additionally, dimensionality reduction techniques, such as principal component analysis, could be applied to the model parameters. This reduction would impact only the machine learning stage, while localization coefficients could still be computed in the full parameter space. Both alternatives will be explored in future work.

For the correction of the prior covariance, a limitation is the need of knowing and storing the actual prior auto-covariance of the model parameters. For scalar model parameters the prior is typically prescribed and because its number is often limited to few hundreds, it is feasible to store the full matrix; however, for grid parameters where there are a large number of gridblocks the prior auto-covariance is usually not explicitly construct, as a results, the proposed correction cannot be applied.     

For future investigation, in addition to strategies aimed at reducing computational cost, the effect of the threshold parameter ($\eta$) in Step 5 of Algorithm~\ref{algo:method.ml-local} should be examined. Increasing the threshold may help further preserve the posterior variance of the model parameters, but it could also negatively impact the quality of the data match.

\section{Conclusions}
\label{sec:conc}

This work presented a machine learning-based, distance-free localization method that enhances ensemble-based data assimilation by mitigating variance underestimation. The method is simple to implement and requires no additional simulations or hyperparameter tuning. It improved the accuracy of cross-covariance estimates and preserved ensemble variance without compromising data match quality. A simple analytical correction to the prior cross-covariance was also introduced and further enhanced performance for scalar/low-dimensional problems. Tested on both scalar and grid-based reservoir models, the machine learning-based localization consistently enhanced data assimilation results, yielding higher posterior variance and accurate data matches, while closely approximating results from large reference ensembles.

\section*{Data Availability}

The code and data for the machine learning-based localization (ML-localization) can be found at \url{https://github.com/viluiz/ml_localization}

\section*{Declaration of competing interest}

The authors declare that they have no known competing financial interests or personal relationships that could have appeared to influence the work reported in this paper.

\section*{Acknowledgements}

The authors would like to thank Petrobras for financial support and permission to publish the article.

\bibliography{./refs/A-D,./refs/E-H,./refs/I-L,./refs/M-O,./refs/P-S,./refs/T-Z}
\bibliographystyle{./refs/abbrvnat_mod.bst}


\end{document}